%% file: main.tex
\documentclass[runningheads]{llncs}

\usepackage[mobile]{eccv}

\usepackage{eccvabbrv}

\usepackage{bbding}
\usepackage{booktabs}
\usepackage{graphicx}	
\usepackage{amsmath}	
\usepackage{amssymb}	
\usepackage{booktabs}
\usepackage{microtype}
\usepackage{epsfig}
\usepackage{float}
\usepackage{placeins}
\usepackage{color, colortbl}
\usepackage{stfloats}
\usepackage{enumitem}
\usepackage{tabularx}
\usepackage{xstring}
\usepackage{multirow}
\usepackage{xspace}
\usepackage{url}
\usepackage{subcaption}
\usepackage{xcolor}
\usepackage{algorithm}
\usepackage{algorithmic}
\usepackage{listings}
\usepackage{pifont}
\usepackage{xcolor}
\usepackage{todonotes}
\definecolor{darkgreen}{RGB}{0,100,0}
\definecolor{mediumgreen}{RGB}{0,128,0}
\definecolor{lightgreen}{RGB}{144,238,144}
\definecolor{limegreen}{RGB}{50,205,50}
\definecolor{forestgreen}{RGB}{34,139,34}
\usepackage[accsupp]{axessibility}  

\usepackage{wrapfig}
\usepackage{pifont}

\usepackage[pagebackref,breaklinks,colorlinks]{hyperref}

\usepackage{orcidlink}

\begin{document}

\title{$\text{R}^2$-Bench: Benchmarking the Robustness of Referring Perception Models under Perturbations} 


\author{Xiang Li$^1$, Kai Qiu$^1$, Jinglu Wang$^2$, Xiaohao Xu$^3$, Rita Singh$^1$, Kashu Yamazak$^1$, Hao Chen$^1$, Xiaonan Huang$^3$, Bhiksha Raj$^1$}

\authorrunning{Xiang Li et al.}

\institute{$^1$ Carnegie Mellon University, $^2$ Microsoft Research, $^3$University of Michigan}

\maketitle

\begin{figure}[h!]
    \centering
    \vspace{-0.7cm}\includegraphics[width=\textwidth]{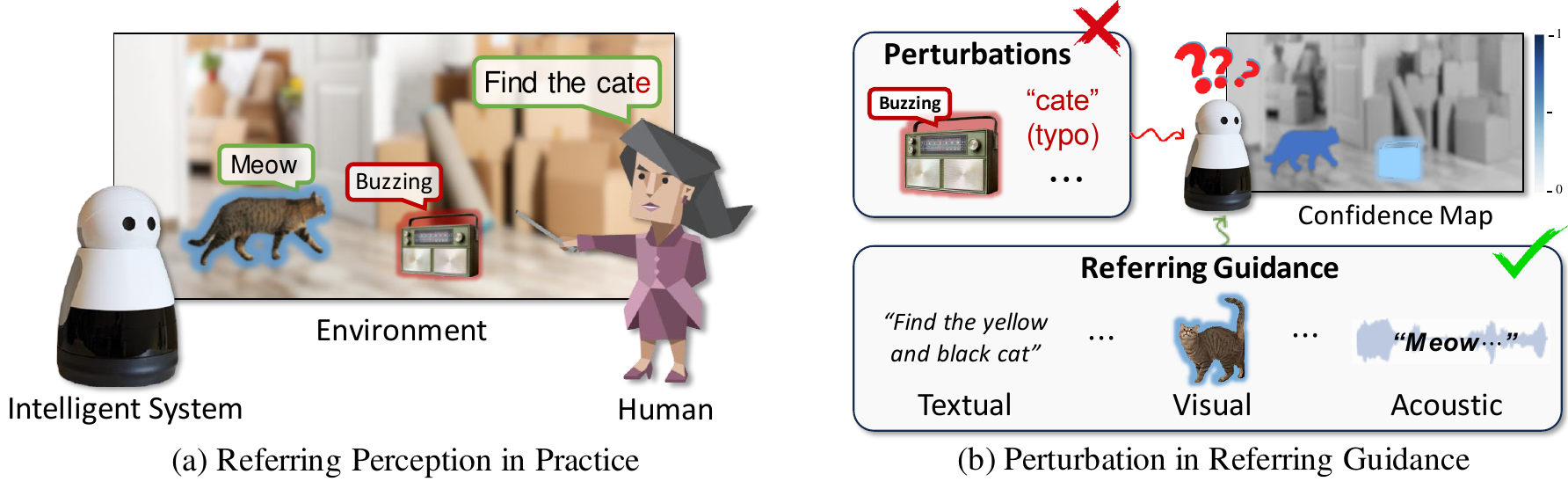}
    \vspace{-0.7cm}
    \caption{
    \textbf{Motivation illustration.} 
    Referring perception models (RPMs) empower intelligent systems with their ability to perform object grounding within the environment based on referring guidance, such as textual descriptions, imagery exemplars, or auditory signals associated with the target object. However, RPMs' performance can be compromised by disturbances in real-world scenarios, such as environmental noise (\eg, extraneous sounds from a nearby radio), human-induced errors (\eg, typographical errors in textual input), and limitations in the sensor (\eg, motion blur in images).
    Conducting a rigorous analysis of RPMs' robustness to a wide array of perturbations is necessary for building reliable real-world applications. 
    }
    \label{fig:teaser}
    \vspace{-1.cm}
\end{figure}

\begin{abstract}

\input{src/0-abstract}
  \vspace{-0.2cm}
  \vspace{-0.2cm}
\end{abstract}

\input{src/tables}
\input{src/supp_tables}

\input{src/1-intro}

\input{src/2-related}

\input{src/3-method}

\input{src/4-experiments}

\input{src/5-conclusion}

\clearpage
\bibliographystyle{splncs04}
\bibliography{main.bib}

\clearpage

\end{document}

%% file: src/0-abstract.tex
Referring perception, which aims at grounding visual objects with multimodal referring guidance, is essential for bridging the gap between humans, who provide instructions, and the environment where intelligent systems perceive. 
Despite progress in this field, the robustness of referring perception models (RPMs) against disruptive perturbations is not well explored. This work thoroughly assesses the resilience of RPMs against various perturbations in both general and specific contexts. 
Recognizing the complex nature of referring perception tasks, we present a comprehensive taxonomy of perturbations, and then develop a versatile toolbox for synthesizing and evaluating the effects of composite disturbances. Employing this toolbox, we construct \textbf{$\text{R}^2$-Bench}, a benchmark for assessing the \textbf{R}obustness of \textbf{R}eferring perception models under noisy conditions across five key tasks. Moreover, we propose the \textbf{$\text{R}^2$-Agent}, an LLM-based agent that simplifies and automates model evaluation via natural language instructions. Our investigation uncovers the vulnerabilities of current RPMs to various perturbations and provides tools for assessing model robustness, potentially promoting the safe and resilient integration of intelligent systems into complex real-world scenarios.

%% file: src/tables.tex
\def\tabris#1{
\begin{table}[#1]
\centering
\vspace{-0.1cm}
\caption{Performance of referring image segmentation methods under low, medium, and high perturbation levels. Average performance change (APC) is averaged among all perturbation levels. We report mIoU following convention.}
\vspace{-0.3cm}
\scalebox{0.69}{
\begin{tabular}{l|p{1.65cm}<{\centering}|p{0.8cm}<{\centering}|p{0.8cm}<{\centering}p{0.8cm}<{\centering}p{0.8cm}<{\centering}|p{0.8cm}<{\centering}|p{0.8cm}<{\centering}|p{0.8cm}<{\centering}p{0.8cm}<{\centering}p{0.8cm}<{\centering}|p{0.8cm}<{\centering}|p{0.8cm}<{\centering}|p{0.8cm}<{\centering}p{0.8cm}<{\centering}p{0.8cm}<{\centering}|p{0.8cm}<{\centering}}
\hline
\multirow{2}*{Method} & \multirow{2}*{Venue} & \multicolumn{5}{c|}{RefCOCO} &
\multicolumn{5}{c|}{RefCOCO+} & \multicolumn{5}{c}{RefCOCOg} \\
\cline{3-17}
~ & ~ & Clean & Low & Med. & High & APC & Clean & Low & Med. & High & APC & Clean & Low & Med.& High & APC \\ 
\hline
LAVT \cite{yang2022lavt} & CVPR 22 & 0.74 & 0.58 & 0.53 & 0.48 & -0.21 & 0.66 & 0.58 & 0.51 & 0.43 & -0.15 &0.63 & 0.55 & 0.47 & 0.38 & -0.16\\
PolyFormer \cite{liu2023polyformer}& CVPR 23 & \bf0.76 & \bf0.70 & \bf0.64 & \bf0.57 & \bf-0.12 & \bf0.71 & \bf0.65 & \bf0.57 & \bf0.50 & -0.14 & \bf0.69 & \bf0.64 & \bf0.57 & \bf0.50 & \bf-0.12 \\
X-Decoder \cite{zou2023generalized} & CVPR 23 & 0.68 & 0.60 & 0.52 & 0.44 &-0.16& 0.69 & 0.60 & 0.51 & 0.44 & -0.17& 0.64 & 0.58 & 0.50 & 0.44 & -0.14\\
ETRIS \cite{xu2023bridging} & ICCV 23 & 0.71 & 0.64 & 0.57 & 0.49 & -0.14 & 0.60 & 0.56 & 0.48 & 0.40 & \bf-0.12 & 0.60 & 0.53 & 0.45 & 0.38 & -0.15\\
SEEM \cite{zou2024segment} & NeurIPS 23 & 0.65 & 0.59 & 0.54 & 0.47 & \bf-0.12 & 0.64 & 0.58 & 0.51 & 0.44 & -0.13 & 0.62 & 0.56 & 0.49 & 0.43 & \bf-0.12\\
\hline
\end{tabular}}
\label{tab:ris}
\vspace{-0.3cm}
\end{table}
}

\def\tabvos#1{
\begin{table}[#1]
\centering
\caption{Performance of video object segmentation methods under low, medium, and high perturbation levels. APC is averaged among all perturbation levels.}
\vspace{-0.3cm}
\scalebox{0.68}{
\begin{tabular}{l|p{1.65cm}<{\centering}|p{0.6cm}<{\centering}p{0.6cm}<{\centering}|p{0.6cm}<{\centering}p{0.6cm}<{\centering}p{0.6cm}<{\centering}p{0.6cm}<{\centering}p{0.6cm}<{\centering}p{0.6cm}<{\centering}|p{0.63cm}<{\centering}p{0.63cm}<{\centering}|p{0.6cm}<{\centering}p{0.6cm}<{\centering}|p{0.6cm}<{\centering}p{0.6cm}<{\centering}p{0.6cm}<{\centering}p{0.6cm}<{\centering}p{0.6cm}<{\centering}p{0.6cm}<{\centering}|p{0.63cm}<{\centering}p{0.63cm}<{\centering}}
\hline
\multirow{3}*{Method} & \multirow{3}*{Venue} & \multicolumn{10}{c|}{DAVIS} &
\multicolumn{10}{c}{YTVOS} \\
\cline{3-22}
~ & ~ & \multicolumn{2}{c|}{Clean} &\multicolumn{2}{c}{Low} & \multicolumn{2}{c}{Medium} & \multicolumn{2}{c|}{High} & \multicolumn{2}{c|}{APC} & \multicolumn{2}{c|}{Clean} & \multicolumn{2}{c}{Low} & \multicolumn{2}{c}{Medium} & \multicolumn{2}{c|}{High} & \multicolumn{2}{c}{APC}\\
\cline{3-22}
~& ~ & $\mathcal{J}$ & $\mathcal{F}$ & $\mathcal{J}$ & $\mathcal{F}$ & $\mathcal{J}$ & $\mathcal{F}$ & $\mathcal{J}$ & $\mathcal{F}$ & $\mathcal{J}$ & $\mathcal{F}$ & $\mathcal{G}_s$ & $\mathcal{G}_u$ & $\mathcal{G}_s$ & $\mathcal{G}_u$ &  $\mathcal{G}_s$ & $\mathcal{G}_u$ &  $\mathcal{G}_s$ & $\mathcal{G}_u$ &  $\mathcal{G}_s$ & $\mathcal{G}_u$\\
\hline
AOT \cite{yang2021associating}& NeurIPS 21 & 76.0 & 83.0 & 71.7 & 79.2 & 69.7 & 76.9 & 65.3 & 71.1 & -7.1 & -7.3 & 86.4 & 84.2 & 84.9 & 80.9 & 82.0 & 78.1 & 76.1 & 70.4 & -5.4 & -7.7 \\
DeAOT\cite{yang2022decoupling} & NeurIPS 22 & 76.9 & 84.5 &  73.2 & 81.0 &72.4 & 80.4 & 68.2 & 75.5 & -5.6 & -6.4 & 86.7 & \bf84.5 & 86.6 & \bf83.7 & 84.5 & \bf81.8 & \bf80.8 & \bf76.8 & \bf-2.7 & \bf-3.7 \\
XMem \cite{cheng2022xmem} & ECCV 22 & 77.4 & 84.5 & 72.2 & 80.0 & 71.8 & 79.7 & 68.4 & 74.7 & -6.6 & -6.4 & 86.5 & \bf84.5 & 84.0 & 81.2 & 81.7 & 79.0 & 78.6 & 74.9 & -5.1 & -6.1\\
DEVA\cite{cheng2023tracking} & ICCV 23 & 78.7 & 85.9 & \bf76.4 & \bf84.9 & 74.2 & 82.4 & \bf69.9 & \bf\bf77.5 & \bf-5.2 & \bf-4.3 & 87.2 & 83.9 & \bf86.7 & 82.8 & \bf84.7 & 80.0 & 80.6 & 76.3 & -3.2 & -4.2\\
Cutie \cite{cheng2023putting} & CVPR 24 & \bf80.6 & \bf87.7 & 75.8 & 83.5 & \bf75.0 & \bf82.7 & 69.3 & 75.4 & -7.2 & -7.2 & \bf87.8 & \bf84.5 & 86.5 & 82.8 & 84.2 & 80.5 & 80.3 & 75.1 & -4.1 & -5.0\\
\hline
\end{tabular}}
\label{tab:vos}
\vspace{-0.3cm}
\end{table}
}

\def\tabrvosdyn#1{
\begin{table}[#1]
\centering
\caption{Robustness of referring video object segmentation methods under none, static and dynamic perturbations.}
\vspace{-0.3cm}
\scalebox{0.76}{
\begin{tabular}{l|p{0.72cm}<{\centering}p{0.72cm}<{\centering}|p{1.2cm}<{\centering}p{1.2cm}<{\centering}|p{1.2cm}<{\centering}p{1.2cm}<{\centering}}
\hline
\multirow{3}*{Method} &
\multicolumn{6}{c}{Ref-YTVOS} \\
\cline{2-7}
~ & \multicolumn{2}{c|}{Clean} & \multicolumn{2}{c|}{Static} & \multicolumn{2}{c}{Dynamic} \\ 
\cline{2-7}
~ & $\mathcal{J}$ & $\mathcal{F}$ & $\mathcal{J}$ & $\mathcal{F}$ & $\mathcal{J}$ & $\mathcal{F}$ \\
\hline
MTTR & 54.0 & 56.4 & 43.0$_{\color{red}{-9.0}}$ & 45.5$_{\color{red}{-10.9}}$ & 44.5$_{\color{red}{-9.5}}$ & 47.4$_{\color{red}{-9.0}}$ \\
ReferFormer & 54.8 & 56.5 & 41.1$_{\color{red}{-13.7}}$ & 37.8$_{\color{red}{-18.7}}$ & 37.4$_{\color{red}{-17.8}}$ & 37.9$_{\color{red}{-18.6}}$\\
$\text{R}^2$-VOS & 56.1 & 58.4 & 45.3$_{\color{red}{-10.8}}$ & 47.4$_{\color{red}{-11.0}}$ & 44.9$_{\color{red}{-11.2}}$ & 46.8$_{\color{red}{-11.6}}$ \\
OnlineRefer & 55.6 & 58.9 & 47.6$_{\color{red}{-8.0}}$ & 49.8$_{\color{red}{-9.1}}$ & 46.3$_{\color{red}{-9.3}}$ & 48.6$_{\color{red}{-10.3}}$ \\
SgMg & 57.7 & 60.0 & 52.4$_{\color{red}{-5.3}}$ & 54.5$_{\color{red}{-5.5}}$ & 51.9$_{\color{red}{-5.8}}$ & 54.3$_{\color{red}{-5.7}}$ \\
\hline
\end{tabular}}
\end{table}
}

\def\tabvosdyn#1{
\begin{table}[#1]
\centering
\caption{Robustness of video object segmentation methods under low, medium, and high perturbations. The subscript $_s$ and $_u$ denote seen and unseen object categories respectively.}
\vspace{-0.3cm}
\scalebox{0.76}{
\begin{tabular}{l|p{0.72cm}<{\centering}p{0.72cm}<{\centering}p{0.72cm}<{\centering}p{0.72cm}<{\centering}|p{1.2cm}<{\centering}p{1.2cm}<{\centering}p{1.2cm}<{\centering}p{1.2cm}<{\centering}|p{1.2cm}<{\centering}p{1.2cm}<{\centering}p{1.2cm}<{\centering}p{1.2cm}<{\centering}}
\hline
\multirow{3}*{Method} &
\multicolumn{12}{c}{YTVOS} \\
\cline{2-13}
~ & \multicolumn{4}{c|}{Clean} & \multicolumn{4}{c|}{Static} & \multicolumn{4}{c}{Dynamic} \\ 
\cline{2-13}
~ & $\mathcal{J}_s$ & $\mathcal{F}_s$ & $\mathcal{J}_u$ & $\mathcal{F}_u$ & $\mathcal{J}_s$ & $\mathcal{F}_s$& $\mathcal{J}_u$ & $\mathcal{F}_u$& $\mathcal{J}_u$ & $\mathcal{F}_s$ & $\mathcal{J}_u$ & $\mathcal{F}_u$ \\
\hline
AOT\cite{yang2021associating} & 83.9 & 88.8 & 79.9 & 88.5 & 79.0$_{\color{red}{-4.9}}$ & 83.1$_{\color{red}{-5.7}}$ & 72.4$_{\color{red}{-7.5}}$ & 80.5$_{\color{red}{-8.0}}$ & 77.2$_{\color{red}{-6.7}}$ & 81.0$_{\color{red}{-7.8}}$ & 68.9$_{\color{red}{-11.0}}$ & 76.7$_{\color{red}{-11.8}}$ \\
DEAOT\cite{yang2022decoupling} & 84.2 & 89.2 & 80.2 & \bf88.8 & 81.6$_{\bf\color{red}{-2.6}}$ & 85.7$_{\color{red}{-3.5}}$ & 75.5$_{\color{red}{-4.7}}$ & 83.5$_{\color{red}{-5.0}}$ & \bf81.4$_{\bf\color{red}{-2.8}}$ & \bf86.2$_{\bf\color{red}{-3.0}}$ & \bf74.4$_{\bf\color{red}{-5.8}}$ & \bf82.4$_{\bf\color{red}{-6.4}}$\\
XMem\cite{cheng2022xmem} & 84.3 & 88.6 & 80.3 & 88.6& 79.5$_{\color{red}{-4.8}}$ & 83.4$_{\color{red}{-5.2}}$ & 74.4$_{\color{red}{-5.9}}$ & 82.3$_{\color{red}{-6.3}}$ & 77.5$_{\color{red}{-6.8}}$ & 81.2$_{\color{red}{-7.4}}$ & 70.9$_{\color{red}{-9.4}}$ & 78.0$_{\color{red}{-10.6}}$\\
DEVA\cite{cheng2023tracking} & 85.0 & 89.4 & 79.7 & 88.0 & \bf81.8$_{\color{red}{-3.2}}$ & \bf86.1$_{\bf\color{red}{-3.3}}$ & \bf75.7$_{\bf\color{red}{-4.0}}$ & \bf83.7$_{\bf\color{red}{-4.3}}$ & 79.8$_{\color{red}{-5.2}}$ & 83.8$_{\color{red}{-5.6}}$ & 71.6$_{\color{red}{-8.1}}$ & 79.2$_{\color{red}{-8.8}}$\\
Cutie\cite{cheng2023putting} & \bf85.6 & \bf90.0 & \bf80.6 & 88.3 & 81.6$_{\color{red}{-4.0}}$ & 85.7$_{\color{red}{-4.3}}$ & 75.5$_{\color{red}{-5.1}}$ & 83.5$_{\color{red}{-4.7}}$ & 79.7$_{\color{red}{-5.9}}$ & 83.7$_{\color{red}{-6.3}}$ & 71.6$_{\color{red}{-9.0}}$ & 79.0$_{\color{red}{-9.3}}$ \\
\hline
\end{tabular}}
\end{table}
}

\def\tabrvos#1{
\begin{table}[#1]
\centering
\caption{Performance of referring video object segmentation methods under low, medium, and high perturbation levels. APC is averaged among all perturbation levels.}
\vspace{-0.3cm}
\scalebox{0.63}{
\begin{tabular}{l|p{1.4cm}<{\centering}|p{0.65cm}<{\centering}p{0.65cm}<{\centering}|p{0.65cm}<{\centering}p{0.65cm}<{\centering}|p{0.65cm}<{\centering}p{0.65cm}<{\centering}|p{0.65cm}<{\centering}p{0.65cm}<{\centering}|p{0.7cm}<{\centering}p{0.8cm}<{\centering}|p{0.65cm}<{\centering}p{0.65cm}<{\centering}|p{0.65cm}<{\centering}p{0.65cm}<{\centering}|p{0.65cm}<{\centering}p{0.65cm}<{\centering}|p{0.65cm}<{\centering}p{0.65cm}<{\centering}|p{0.7cm}<{\centering}p{0.7cm}<{\centering}}
\hline
\multirow{3}*{Method} & \multirow{3}*{Venue} & \multicolumn{10}{c|}{Ref-DAVIS} &
\multicolumn{10}{c}{Ref-YTVOS} \\
\cline{3-22}
~ & ~ & \multicolumn{2}{c|}{Clean} &\multicolumn{2}{c|}{Low} & \multicolumn{2}{c|}{Medium} & \multicolumn{2}{c|}{High} & \multicolumn{2}{c|}{APC} & \multicolumn{2}{c|}{Clean} & \multicolumn{2}{c|}{Low} & \multicolumn{2}{c|}{Medium} & \multicolumn{2}{c|}{High} & \multicolumn{2}{c}{APC} \\ 
\cline{3-22}
~& ~ & $\mathcal{J}$ & $\mathcal{F}$ & $\mathcal{J}$ & $\mathcal{F}$ & $\mathcal{J}$ & $\mathcal{F}$ & $\mathcal{J}$ & $\mathcal{F}$ & $\mathcal{J}$ & $\mathcal{F}$ & $\mathcal{J}$ & $\mathcal{F}$ & $\mathcal{F}$ & $\mathcal{J}$ &  $\mathcal{J}$ & $\mathcal{F}$ &  $\mathcal{J}$ & $\mathcal{F}$ & $\mathcal{J}$ & $\mathcal{F}$\\
\hline
\multicolumn{22}{c}{Video-level Backbone} \\
\hline
MTTR \cite{botach2021mttr} & CVPR 22 & - & - & - & - & - & - & - & - & - & - & 54.0 & 56.4 & 48.9 & 51.6 & 43.0 & 45.5 & 38.0 & 40.7 & -10.7 & -10.5 \\
SgMg \cite{miao2023spectrum} & ICCV 23 & \bf59.0 & \bf64.8 & \bf57.3 & \bf62.6 & \bf51.9 & \bf57.7 & \bf47.3 & \bf51.1 & \bf-6.8 & \bf-7.7 & \bf57.7 & \bf60.0 & \bf57.4 & \bf59.9 & \bf52.4 & \bf54.5 & \bf44.1 & \bf46.2 & \bf-6.4 & \bf-6.5\\
\hline
\multicolumn{22}{c}{Image-level Backbone} \\
\hline
ReferFormer \cite{referformer} & CVPR 22 & 55.8 & 61.3 & 46.5 & 50.1 & 43.1 & 48.0 & 38.5 & 41.9 & -13.1 & -14.6 & 54.8 & 56.5 & 41.1 & 42.0 & 37.4 & 37.8 & 33.7 & 34.2 & -17.4 & -18.5 \\
OnlineRefer \cite{wu2023onlinerefer} & ICCV 23 & 55.7 & \bf62.9 & \bf51.6 & \bf57.9 & \bf48.0 & \bf54.3 & \bf42.2 & \bf46.3 & \bf-8.4 & \bf-10.0 & 55.6 & \bf58.9 & \bf51.5 & \bf54.3 & \bf47.6 & \bf49.8 & \bf39.6 & \bf42.0 & \bf-9.4 & -10.2 \\
$\text{R}^2$-VOS \cite{li2023robust} & ICCV 23 & \bf57.2 & 62.4 & 50.2 & 57.0 & 46.5 & 53.8 & 39.2 & 44.6 & -11.9 & -10.6 & \bf56.1 & 58.4 & 48.4 & 51.2 & 45.3 & 47.4 & 38.5 & 40.7 & -12.0 & \bf-9.7 \\
\hline
\end{tabular}}
\label{tab:rvos}
\vspace{-0.3cm}
\end{table}
}

\def\tabavs#1{
\begin{table}[#1]
\centering
\caption{Performance of audiovisual segmentation methods under low, medium, and high perturbation levels. APC is averaged among all perturbation levels.}
\vspace{-0.3cm}
\scalebox{0.65}{
\begin{tabular}{l|p{1.4cm}<{\centering}|p{0.6cm}<{\centering}p{0.6cm}<{\centering}|p{0.6cm}<{\centering}p{0.6cm}<{\centering}|p{0.6cm}<{\centering}p{0.6cm}<{\centering}|p{0.6cm}<{\centering}p{0.6cm}<{\centering}|p{0.7cm}<{\centering}p{0.7cm}<{\centering}|p{0.6cm}<{\centering}p{0.6cm}<{\centering}|p{0.6cm}<{\centering}p{0.6cm}<{\centering}|p{0.6cm}<{\centering}p{0.6cm}<{\centering}|p{0.6cm}<{\centering}p{0.6cm}<{\centering}|p{0.7cm}<{\centering}p{0.7cm}<{\centering}}
\hline
\multirow{3}*{Method} & \multirow{3}*{Venue} & \multicolumn{10}{c|}{AVS-s4} &
\multicolumn{10}{c}{AVS-ms3} \\
\cline{3-22}
~ & ~ & \multicolumn{2}{c|}{Clean} &\multicolumn{2}{c|}{Low} & \multicolumn{2}{c|}{Medium} & \multicolumn{2}{c|}{High} & \multicolumn{2}{c|}{APC} & \multicolumn{2}{c|}{Clean} & \multicolumn{2}{c|}{Low} & \multicolumn{2}{c|}{Medium} & \multicolumn{2}{c|}{High} & \multicolumn{2}{c}{APC} \\ 
\cline{3-22}
~ & ~ & $\mathcal{J}$ & $\mathcal{F}$ & $\mathcal{J}$ & $\mathcal{F}$ & $\mathcal{J}$ & $\mathcal{F}$ & $\mathcal{J}$ & $\mathcal{F}$ & $\mathcal{J}$ & $\mathcal{F}$ & $\mathcal{J}$ & $\mathcal{F}$ & $\mathcal{F}$ & $\mathcal{J}$ &  $\mathcal{J}$ & $\mathcal{F}$ &  $\mathcal{J}$ & $\mathcal{F}$ &  $\mathcal{J}$ & $\mathcal{F}$ \\
\hline
AVS \cite{zhou2022avs} & ECCV 22 & 72.8 & 84.8 & 68.4 & 79.7 & 60.8 & 71.2 & 55.6 & 66.8 & -11.2 & -12.2 & 47.9 & 57.8 & 44.4 & 54.6 & 41.2 & 50.5 & 37.0 & 44.2 & -7.0 & \bf-8.0\\
CATR \cite{li2023catr} & MM 23 & 74.9 & \bf87.1 & 71.1 & \bf84.2 & 65.7 & \bf79.0 & 58.4 & \bf72.4 & \bf-9.8 & -8.6 &53.1 & \bf65.6 & 49.8 & \bf61.4 & 46.7 & \bf58.4 & 41.2 & \bf52.9 & -7.2 &\bf-8.0\\
AVSegFormer \cite{gao2023avsegformer} & AAAI 23 & 76.4 & 86.7 & 70.9 & 82.3 & 62.4 & 74.8 & 57.3 & 69.8 & -12.9 & -11.0 &  49.5 & 62.8 & 46.9 & 59.0 & 43.7 & 55.4 & 39.8 & 50.1 & \bf-6.0 & \bf-8.0\\
QSD \cite{li2023towards} & CVPR 24 & \bf77.6 & 85.6 & \bf74.8 & 83.6 & \bf68.3 & 78.4 & \bf59.4 & 71.2 & -10.1 & \bf-7.9&  \bf61.8 & 64.3 & \bf56.6 & 60.0 & \bf53.1 & 57.1 & \bf47.6 & 51.6 & -9.4 & \bf-8.0\\
\hline
\end{tabular}}
\label{tab:avs}
\vspace{-0.3cm}
\end{table}
}

\def\tabrvosP#1{
\begin{table}[#1]
\centering
\caption{Robustness of referring video object segmentation methods under different textual and visual perturbations on AVS-ms3.}
\vspace{-0.3cm}
\begin{tabular}{l|p{0.72cm}<{\centering}p{0.72cm}<{\centering}|p{0.72cm}<{\centering}p{0.72cm}<{\centering}p{0.72cm}<{\centering}p{0.72cm}<{\centering}|p{0.72cm}<{\centering}p{0.72cm}<{\centering}|p{0.72cm}<{\centering}p{0.72cm}<{\centering}p{0.72cm}<{\centering}p{0.72cm}<{\centering}p{0.72cm}<{\centering}p{0.72cm}<{\centering}p{0.72cm}<{\centering}p{0.72cm}<{\centering}p{0.72cm}<{\centering}}
\hline
\multirow{2}*{Method} & \multicolumn{8}{c|}{Visual Perturbations} & \multicolumn{4}{c}{Textual Perturbations} \\ 
\cline{2-13}
~& Snow & Fog & DB & GB & IN & SN & JPEG & PIX & MS & PE & GE & MW \\
\hline
ReferFormer & 44.8 & 48.1 & 43.5 & 44.6 & 44.9 & 45.3 & 50.0 & 45.5 & 47.8 & 49.4 & 49.3 & 46.5 \\
$\text{R}^2$-VOS & 46.8 & 51.2 & 48.0 & 49.9 & 44.0 & 43.6 & 55.6 & 51.5 & 52.2 & 52.6 & 51.8 & 51.1 \\
OnlineRefer & 44.5 & 48.1 & 43.5 & 44.6 & 44.9 & 45.3 & 50.0 & 45.5 & 47.8 & 49.4 &  49.7 & 46.5 \\
SgMg & 54.6 & 56.9 & 55.5 & 57.3 & 53.3 & 52.9 & 59.8 & 56.7 & 57.7 & 58.2 & 56.3 & 53.0 \\
\hline
\end{tabular}
\end{table}
\label{tab:perturb type davis}
}

\def\tabavsP#1{
\begin{table}[#1]
\centering
\caption{Robustness of audiovisual segmentation methods under different acoustic perturbations on AVS-s4.}
\vspace{-0.3cm}
\begin{tabular}{l|p{0.72cm}<{\centering}p{0.72cm}<{\centering}p{0.72cm}<{\centering}p{0.72cm}<{\centering}p{0.72cm}<{\centering}p{0.72cm}<{\centering}p{0.72cm}<{\centering}p{0.72cm}<{\centering}p{0.72cm}<{\centering}p{0.72cm}<{\centering}p{0.72cm}<{\centering}p{0.72cm}<{\centering}p{0.72cm}<{\centering}p{0.72cm}<{\centering}p{0.72cm}<{\centering}p{0.72cm}<{\centering}p{0.72cm}<{\centering}}
\hline
\multirow{2}*{Method} & \multicolumn{12}{c}{Acoustic Perturbations} \\ 
\cline{2-13}
~& GA & MP3 & RS & AA & PF & LP & HP & IR & TM & GN & TD & BN \\
\hline
AVS \\
CATR \\
AVSegFormer & 81.6 & 81.6 & 81.5 & 81.6 & 81.5 & 81.5 & 81.5 & 81.3 & 81.4 & 81.5 & 81.6 & 81.5 \\
QSD & 81.7 & 81.7 & 80.9 & 81.5 & 81.5 & 80.9 & 80.9 & 79.1 & 81.3 & 81.7 & 81.5 & 81.4 \\
\hline
\end{tabular}
\end{table}
}

\def\tabdata#1{
\begin{wraptable}{#1}{6.2cm}
\centering
\vspace{-1.1cm}
\caption{Data selection performance.}
\scalebox{0.72}{
\begin{tabular}{l|p{0.7cm}<{\centering}p{0.7cm}<{\centering}|p{0.7cm}<{\centering}p{0.7cm}<{\centering}|p{0.7cm}<{\centering}p{0.7cm}<{\centering}|p{0.7cm}<{\centering}p{0.7cm}<{\centering}}
\hline
\# Ver. Iter. & \multicolumn{2}{c|}{\XSolidBrush} & \multicolumn{2}{c|}{1} & \multicolumn{2}{c|}{2} & \multicolumn{2}{c}{3} \\
\hline
Score & ACC & Rate & ACC & Rate & ACC & Rate & ACC & Rate \\
\hline
Data Sample & 0.53 & 0.78 & 0.64 & 0.81 & 0.77 & 0.85 & 0.80 & 0.91 \\
Perturbation & 0.65 & 0.87 & 0.77 & 0.86 & 0.86 & 0.91 & 0.87 & 0.93 \\
Metric Func. & 1.00 & 1.00 & 1.00 & 1.00 & 1.00 & 1.00 & 1.00 & 1.00 \\
\hline
\hline
\end{tabular}}
\label{tab:data}
\vspace{-0.8cm}
\end{wraptable}
}

%% file: src/supp_tables.tex
\def\tabdavissupp#1{
\begin{table}[#1]
\centering
\scalebox{0.82}{
\begin{tabular}{l|p{0.72cm}<{\centering}p{0.72cm}<{\centering}|p{0.72cm}<{\centering}p{0.72cm}<{\centering}|p{0.72cm}<{\centering}p{0.72cm}<{\centering}|p{0.72cm}<{\centering}p{0.72cm}<{\centering}|p{0.72cm}<{\centering}p{0.72cm}<{\centering}|p{0.72cm}<{\centering}p{0.72cm}<{\centering}|p{0.72cm}<{\centering}p{0.72cm}<{\centering}|p{0.72cm}<{\centering}p{0.72cm}<{\centering}}
\hline
\multirow{3}*{Method} & \multicolumn{8}{c|}{Low} & \multicolumn{8}{c}{Medium} \\ 
\cline{2-17}
~& \multicolumn{2}{c|}{Anno0} & \multicolumn{2}{c|}{Anno1} & \multicolumn{2}{c|}{Anno2} & \multicolumn{2}{c|}{Anno3} & \multicolumn{2}{c|}{Anno0} & \multicolumn{2}{c|}{Anno1} & \multicolumn{2}{c|}{Anno2} & \multicolumn{2}{c}{Anno3} \\
~& $\mathcal{J}$ & $\mathcal{F}$ & $\mathcal{J}$ & $\mathcal{F}$ & $\mathcal{J}$ & $\mathcal{F}$ & $\mathcal{J}$ & $\mathcal{F}$ & $\mathcal{J}$ & $\mathcal{F}$ & $\mathcal{J}$ & $\mathcal{F}$ & $\mathcal{J}$ & $\mathcal{F}$ & $\mathcal{J}$ & $\mathcal{F}$ \\
\hline
ReferFormer & 47.9 & 52.0 & 47.5 & 51.4 & 45.5 & 49.1 & 45.2 & 49.7 & 43.6 & 48.4 & 44.3 & 49.2 & 42.6 & 47.4 & 41.8 & 47.1  \\
$\text{R}^2$-VOS & 50.3 & 57.1 & 49.4 & 56.6 & 49.3 & 56.6 & 51.7 & 57.5 & 47.7 & 55.2 & 47.7 & 55.7 & 44.3 & 51.5 & 46.2 & 52.6 \\
OnlineRefer & 51.4 & 57.7 & 50.4 & 56.8 & 52.5 & 58.9 & 52.2 & 58.2 & 48.0 & 54.2 & 50.3 & 56.1 & 46.9 & 53.6 & 46.9 & 53.5 \\
SgMg & 58.2 & 62.9 & 57.0 & 62.9 & 57/5 & 62.8 & 56.5 & 61.6 & 51.3 & 57.3 & 55.3 & 61.0 & 49.9 & 55.1 & 51.0 & 57.2 \\
\hline
\multirow{3}*{Method} & \multicolumn{8}{c|}{High} & \multicolumn{8}{c}{Dynamic} \\ 
\cline{2-17}
~& \multicolumn{2}{c|}{Anno0} & \multicolumn{2}{c|}{Anno1} & \multicolumn{2}{c|}{Anno2} & \multicolumn{2}{c|}{Anno3} & \multicolumn{2}{c|}{Anno0} & \multicolumn{2}{c|}{Anno1} & \multicolumn{2}{c|}{Anno2} & \multicolumn{2}{c}{Anno3} \\
~& $\mathcal{J}$ & $\mathcal{F}$ & $\mathcal{J}$ & $\mathcal{F}$ & $\mathcal{J}$ & $\mathcal{F}$ & $\mathcal{J}$ & $\mathcal{F}$ & $\mathcal{J}$ & $\mathcal{F}$ & $\mathcal{J}$ & $\mathcal{F}$ & $\mathcal{J}$ & $\mathcal{F}$ & $\mathcal{J}$ & $\mathcal{F}$ \\
\hline
ReferFormer  & 39.4 & 43.3 & 39.0 & 42.6 & 37.8 & 41.0 & 37.7 & 40.6 & 43.0 & 47.3 & 44.1 & 48.0 & 43.0 & 47.0 & 41.4 & 46.3 \\
$\text{R}^2$-VOS & 38.5 & 43.7 & 38.5 & 44.2 & 39.2 & 45.3 & 40.7 & 45.2 & 46.8 & 52.1 & 46.0 & 52.2 & 43.5 & 50.5 & 49.6 & 55.0\\
OnlineRefer & 39.1 & 43.3 & 43.4 & 47.6 & 42.9 & 47.2 & 43.5 & 47.2 & 50.5 & 56.3 & 49.9 & 55.2 & 49.5 & 54.5 & 44.3 & 50.4 \\
SgMg & 47.8 & 51.5 & 48.9 & 51.9 & 46.2 & 50.6 & 46.1 & 50.3 & 53.3 & 58.5 & 53.7 & 58.5 & 54.9 & 60.2 & 51.4 & 56.3 \\
\hline
\end{tabular}}
\caption{Robustness of referring video object segmentation methods under low (L), medium (M) and high (H) perturbation levels.}
\end{table}
}

\def\tabrisrefcocoevalsupp#1{
\begin{table}[#1]
\centering
\scalebox{0.7}{
\begin{tabular}{l|p{0.77cm}<{\centering}p{0.77cm}<{\centering}p{0.77cm}<{\centering}p{0.77cm}<{\centering}p{0.77cm}<{\centering}p{0.77cm}<{\centering}|p{0.77cm}<{\centering}p{0.77cm}<{\centering}p{0.77cm}<{\centering}p{0.77cm}<{\centering}p{0.77cm}<{\centering}p{0.77cm}<{\centering}|p{0.77cm}<{\centering}p{0.77cm}<{\centering}p{0.77cm}<{\centering}p{0.77cm}
<{\centering}p{0.77cm}<{\centering}p{0.77cm}
<{\centering}}
\hline
\multirow{2}*{Method} & \multicolumn{6}{c|}{L} & \multicolumn{6}{c|}{M} & \multicolumn{6}{c}{H}  \\ 
\cline{2-19}
~& mIoU & P@50 & P@60 & P@70 & P@80 & P@90 & mIoU & P@50 & P@60 & P@70 & P@80 & P@90 & mIoU & P@50 & P@60 & P@70 & P@80 &
P@90\\
\hline
LAVT \cite{yang2022lavt}& 0.58 & 0.65 & 0.61 & 0.55 & 0.46 & 0.22 & 0.51 & 0.57 & 0.53 & 0.48 & 0.37 & 0.17 & 0.44 & 0.49 & 0.45 & 0.39 & 0.30 & 0.13 \\
PolyFormer \cite{liu2023polyformer}& 0.70 & 0.82 & 0.78 & 0.71 & 0.57 & 0.21 & 0.64 & 0.75 & 0.71 & 0.64 & 0.50 & 0.18 & 0.57 & 0.66 & 0.62 & 0.55 & 0.42 & 0.14\\
X-Decoder \cite{zou2023generalized}& 0.60 & 0.69 & 0.66 & 0.61 & 0.51 & 0.24 & 0.52 & 0.61 & 0.58 & 0.53 & 0.44 & 0.19 & 0.44 & 0.51 & 0.48 & 0.44 & 0.35 & 0.15\\
ETRIS \cite{xu2023bridging}& 0.64 & 0.75 & 0.70 & 0.62 & 0.46 & 0.14 & 0.57 & 0.66 & 0.61 & 0.53 & 0.38 & 0.11 & 0.49 & 0.56 & 0.51 & 0.43 & 0.29 & 0.08 \\
SEEM \cite{zou2024segment}& 0.59 & 0.66 & 0.64 & 0.62 & 0.55 & 0.34 & 0.54 & 0.61 & 0.60 & 0.57 & 0.51 & 0.30 & 0.47 & 0.53 & 0.51 & 0.47 & 0.40 & 0.22\\
\hline
\end{tabular}}
\caption{Robustness of referring image segmentation on RefCOCO Validation set with low (L), medium (M) and high (H) perturbation levels.}
\end{table}
}

\def\tabrisrefcocotestasupp#1{
\begin{table}[#1]
\centering
\scalebox{0.7}{
\begin{tabular}{l|p{0.77cm}<{\centering}p{0.77cm}<{\centering}p{0.77cm}<{\centering}p{0.77cm}<{\centering}p{0.77cm}<{\centering}p{0.77cm}<{\centering}|p{0.77cm}<{\centering}p{0.77cm}<{\centering}p{0.77cm}<{\centering}p{0.77cm}<{\centering}p{0.77cm}<{\centering}p{0.77cm}<{\centering}|p{0.77cm}<{\centering}p{0.77cm}<{\centering}p{0.77cm}<{\centering}p{0.77cm}
<{\centering}p{0.77cm}<{\centering}p{0.77cm}
<{\centering}}
\hline
\multirow{2}*{Method} & \multicolumn{6}{c|}{L} & \multicolumn{6}{c|}{M} & \multicolumn{6}{c}{H}  \\ 
\cline{2-19}
~& mIoU & P@50 & P@60 & P@70 & P@80 & P@90 & mIoU & P@50 & P@60 & P@70 & P@80 & P@90 & mIoU & P@50 & P@60 & P@70 & P@80 &
P@90\\
\hline
LAVT \cite{yang2022lavt} & 0.64 & 0.73 & 0.69 & 0.63 & 0.53 & 0.25 & 0.56 & 0.64 & 0.60 & 0.54 & 0.43 & 0.19 & 0.48 & 0.54 & 0.50 & 0.44 & 0.33 & 0.13\\
PolyFormer \cite{liu2023polyformer} & 0.72 & 0.84 & 0.81 & 0.75 & 0.58 & 0.19 & 0.66 & 0.78 & 0.75 & 0.68 & 0.52 & 0.17 & 0.58 & 0.68 & 0.65 & 0.58 & 0.42 & 0.13 \\
X-Decoder \cite{zou2023generalized}& 0.62 & 0.71 & 0.69 & 0.65 & 0.53 & 0.22 & 0.55 & 0.65 & 0.62 & 0.58 & 0.47 & 0.18 & 0.46 & 0.54 & 0.51 & 0.46 & 0.36 & 0.12 \\
ETRIS \cite{xu2023bridging} & 0.68 & 0.80 & 0.76 & 0.68 & 0.51 & 0.14 & 0.61 & 0.71 & 0.67 & 0.59 & 0.43 & 0.11 & 0.52 & 0.61 & 0.56 & 0.47 & 0.31 & 0.07\\
SEEM \cite{zou2024segment}& 0.63 & 0.70 & 0.69 & 0.66 & 0.58 & 0.34 & 0.57 & 0.65 & 0.63 & 0.61 & 0.53 & 0.28 & 0.49 & 0.56 & 0.54 & 0.50 & 0.43 & 0.22 \\
\hline
\end{tabular}}
\caption{Robustness of referring image segmentation on RefCOCO Test-A set with low (L), medium (M) and high (H) perturbation levels.}
\end{table}
}

\def\tabrisrefcocotestbsupp#1{
\begin{table}[#1]
\centering
\scalebox{0.7}{
\begin{tabular}{l|p{0.77cm}<{\centering}p{0.77cm}<{\centering}p{0.77cm}<{\centering}p{0.77cm}<{\centering}p{0.77cm}<{\centering}p{0.77cm}<{\centering}|p{0.77cm}<{\centering}p{0.77cm}<{\centering}p{0.77cm}<{\centering}p{0.77cm}<{\centering}p{0.77cm}<{\centering}p{0.77cm}<{\centering}|p{0.77cm}<{\centering}p{0.77cm}<{\centering}p{0.77cm}<{\centering}p{0.77cm}
<{\centering}p{0.77cm}<{\centering}p{0.77cm}
<{\centering}}
\hline
\multirow{2}*{Method} & \multicolumn{6}{c|}{L} & \multicolumn{6}{c|}{M} & \multicolumn{6}{c}{H}  \\ 
\cline{2-19}
~& mIoU & P@50 & P@60 & P@70 & P@80 & P@90 & mIoU & P@50 & P@60 & P@70 & P@80 & P@90 & mIoU & P@50 & P@60 & P@70 & P@80 &
P@90\\
\hline
LAVT \cite{yang2022lavt}& 0.53 & 0.58 & 0.53 & 0.47 & 0.39 & 0.21 & 0.47 & 0.52 & 0.47 & 0.41 & 0.33 & 0.16 & 0.40 & 0.44 & 0.40 & 0.34 & 0.26 & 0.12\\
PolyFormer \cite{liu2023polyformer} & 0.67 & 0.77 & 0.73 & 0.66 & 0.54 & 0.25 & 0.61 & 0.70 & 0.66 & 0.59 & 0.47 & 0.20 & 0.55 & 0.63 & 0.58 & 0.52 & 0.41 & 0.17 \\
X-Decoder \cite{zou2023generalized}& 0.51 & 0.58 & 0.55 & 0.52 & 0.44 & 0.24 & 0.46 & 0.52 & 0.49 & 0.45 & 0.37 & 0.18 & 0.40 & 0.45 & 0.42 & 0.38 & 0.31 & 0.15\\
ETRIS \cite{xu2023bridging}& 0.60 & 0.68 & 0.62 & 0.54 & 0.41 & 0.16 & 0.52 & 0.59 & 0.53 & 0.44 & 0.32 & 0.11 & 0.46 & 51 & 0.45 & 0.37 & 0.26 & 0.09\\
SEEM \cite{zou2024segment}& 0.50 & 0.56 & 0.54 & 0.52 & 0.47 & 0.31 & 0.47 & 0.52 & 0.50 & 0.48 & 0.42 & 0.26 & 0.41 & 0.45 & 0.43 & 0.40 & 0.35 & 0.22\\
\hline
\end{tabular}}
\caption{Robustness of referring image segmentation on RefCOCO Test-B set with low (L), medium (M) and high (H) perturbation levels.}
\end{table}
}

\def\tabrisrefcocopevalsupp#1{
\begin{table}[#1]
\centering
\scalebox{0.7}{
\begin{tabular}{l|p{0.77cm}<{\centering}p{0.77cm}<{\centering}p{0.77cm}<{\centering}p{0.77cm}<{\centering}p{0.77cm}<{\centering}p{0.77cm}<{\centering}|p{0.77cm}<{\centering}p{0.77cm}<{\centering}p{0.77cm}<{\centering}p{0.77cm}<{\centering}p{0.77cm}<{\centering}p{0.77cm}<{\centering}|p{0.77cm}<{\centering}p{0.77cm}<{\centering}p{0.77cm}<{\centering}p{0.77cm}
<{\centering}p{0.77cm}<{\centering}p{0.77cm}
<{\centering}}
\hline
\multirow{2}*{Method} & \multicolumn{6}{c|}{L} & \multicolumn{6}{c|}{M} & \multicolumn{6}{c}{H}  \\ 
\cline{2-19}
~& mIoU & P@50 & P@60 & P@70 & P@80 & P@90 & mIoU & P@50 & P@60 & P@70 & P@80 & P@90 & mIoU & P@50 & P@60 & P@70 & P@80 &
P@90\\
\hline
LAVT \cite{yang2022lavt}& 0.58 & 0.66 & 0.61 & 0.56 & 0.46 & 0.23 & 0.51 & 0.57 & 0.53 & 0.48 & 0.40 & 0.18 & 0.43 & 0.48 & 0.44 & 0.39 & 0.29 & 0.12\\
PolyFormer \cite{liu2023polyformer}& 0.65 & 0.75 & 0.72 & 0.66 & 0.53 & 0.20 & 0.57 & 0.66 & 0.63 & 0.57 & 0.45 & 0.16 & 0.50 & 0.58 & 0.55 & 0.49 & 0.37 & 0.12\\
X-Decoder \cite{zou2023generalized}& 0.60 & 0.69 & 0.66 & 0.62 & 0.52 & 0.24 & 0.51 & 0.59 & 0.56 & 0.52 & 0.42 & 0.19 & 0.44 & 0.51 & 0.49 & 0.44 & 0.35 & 0.15\\
ETRIS \cite{xu2023bridging}& 0.56 & 0.65 & 0.60 & 0.53 & 0.40 & 0.12 & 0.48 & 0.55 & 0.51 & 0.45 & 0.32 & 0.09 & 0.40 & 0.46 & 0.41 & 0.35 & 0.23 & 0.06 \\
SEEM \cite{zou2024segment}& 0.58 & 0.65 & 0.63 & 0.60 & 0.54 & 0.34 & 0.51 & 0.57 & 0.55 & 0.53 & 0.47 & 0.28 & 0.44 & 0.50 & 0.48 & 0.45 & 0.39 & 0.21\\
\hline
\end{tabular}}
\caption{Robustness of referring image segmentation on RefCOCO+ Validation set with low (L), medium (M) and high (H) perturbation levels.}
\end{table}
}

\def\tabrisrefcocoptestasupp#1{
\begin{table}[#1]
\centering
\scalebox{0.7}{
\begin{tabular}{l|p{0.77cm}<{\centering}p{0.77cm}<{\centering}p{0.77cm}<{\centering}p{0.77cm}<{\centering}p{0.77cm}<{\centering}p{0.77cm}<{\centering}|p{0.77cm}<{\centering}p{0.77cm}<{\centering}p{0.77cm}<{\centering}p{0.77cm}<{\centering}p{0.77cm}<{\centering}p{0.77cm}<{\centering}|p{0.77cm}<{\centering}p{0.77cm}<{\centering}p{0.77cm}<{\centering}p{0.77cm}
<{\centering}p{0.77cm}<{\centering}p{0.77cm}
<{\centering}}
\hline
\multirow{2}*{Method} & \multicolumn{6}{c|}{L} & \multicolumn{6}{c|}{M} & \multicolumn{6}{c}{H}  \\ 
\cline{2-19}
~& mIoU & P@50 & P@60 & P@70 & P@80 & P@90 & mIoU & P@50 & P@60 & P@70 & P@80 & P@90 & mIoU & P@50 & P@60 & P@70 & P@80 &
P@90\\
\hline
LAVT \cite{yang2022lavt} & 0.63 & 0.71 & 0.68 & 0.62 & 0.52 & 0.25 & 0.55 & 0.63 & 0.59 & 0.53 & 0.42 & 0.18 & 0.47 & 0.53 & 0.49 & 0.43 & 0.33 & 0.14 \\
PolyFormer \cite{liu2023polyformer}& 0.68 & 0.79 & 0.76 & 0.70 & 0.56 & 0.19 & 0.60 & 0.71 & 0.68 & 0.62  & 0.48 & 0.15 & 0.53 & 0.62 & 0.59 & 0.53 & 0.40 & 0.12\\
X-Decoder \cite{zou2023generalized}& 0.61 & 0.70 & 0.68 & 0.64 & 0.54 & 0.22 & 0.54 & 0.63 & 0.61 & 0.57 & 0.46 & 0.18 & 0.44 & 0.52 & 0.49 & 0.44 & 0.35 & 0.13 \\
ETRIS \cite{xu2023bridging}& 0.62 & 0.73 & 0.68 & 0.61 & 0.42 & 0.12 & 0.53 & 0.63 & 0.59 & 0.52 & 0.37 & 0.09 & 0.44 & 0.51 & 0.47 & 0.40 & 0.27 & 0.06\\
SEEM \cite{zou2024segment}& 0.62 & 0.70 & 0.69 & 0.66 & 0.59 & 0.34 & 0.56 & 0.63 & 0.62 & 0.59 & 0.53 & 0.29 & 0.48 & 0.54 & 0.53 & 0.50 & 0.43 & 0.22\\
\hline
\end{tabular}}
\caption{Robustness of referring image segmentation on RefCOCO+ Test-A set with low (L), medium (M) and high (H) perturbation levels.}
\end{table}
}

\def\tabrisrefcocoptestbsupp#1{
\begin{table}[#1]
\centering
\scalebox{0.7}{
\begin{tabular}{l|p{0.77cm}<{\centering}p{0.77cm}<{\centering}p{0.77cm}<{\centering}p{0.77cm}<{\centering}p{0.77cm}<{\centering}p{0.77cm}<{\centering}|p{0.77cm}<{\centering}p{0.77cm}<{\centering}p{0.77cm}<{\centering}p{0.77cm}<{\centering}p{0.77cm}<{\centering}p{0.77cm}<{\centering}|p{0.77cm}<{\centering}p{0.77cm}<{\centering}p{0.77cm}<{\centering}p{0.77cm}
<{\centering}p{0.77cm}<{\centering}p{0.77cm}
<{\centering}}
\hline
\multirow{2}*{Method} & \multicolumn{6}{c|}{L} & \multicolumn{6}{c|}{M} & \multicolumn{6}{c}{H}  \\ 
\cline{2-19}
~& mIoU & P@50 & P@60 & P@70 & P@80 & P@90 & mIoU & P@50 & P@60 & P@70 & P@80 & P@90 & mIoU & P@50 & P@60 & P@70 & P@80 &
P@90\\
\hline
LAVT \cite{yang2022lavt}& 0.52 & 0.58 & 0.53 & 0.47 & 0.39 & 0.21 & 0.46 & 0.50 & 0.46 & 0.42 & 0.32 & 0.17 & 0.39 & 0.43 & 0.38 & 0.33 & 0.25 & 0.12 \\
PolyFormer \cite{liu2023polyformer}& 0.58 & 0.65 & 0.61 & 0.56 & 0.45 & 0.21 & 0.52 & 0.59 & 0.55 & 0.50 & 0.40 & 0.18 & 0.46 & 0.52 & 0.48 & 0.43 & 0.33 & 0.14\\
X-Decoder \cite{zou2023generalized}& 0.52 & 0.59 & 0.56 & 0.52 & 0.45 & 0.24 & 0.45 & 0.51 & 0.48 & 0.44 & 0.37 & 0.19 & 0.39 & 0.44 & 0.41 & 0.38 & 0.31 & 0.14\\
ETRIS \cite{xu2023bridging}& 0.47 & 0.52 & 0.47 & 0.41 & 0.29 & 0.11 & 0.40 & 0.44 & 0.39 & 0.33 & 0.24 & 0.09 & 0.34 & 0.37 & 0.31 & 0.25 & 0.17 & 0.06 \\
SEEM \cite{zou2024segment}& 0.49 & 0.54 & 0.52 & 0.50 & 0.49 & 0.29 & 0.44 & 0.48 & 0.47 & 0.44 & 0.39 & 0.25 & 0.40 & 0.44 & 0.42 & 0.39 & 0.34 & 0.20\\
\hline
\end{tabular}}
\caption{Robustness of referring image segmentation on RefCOCO+ Test-B set with low (L), medium (M) and high (H) perturbation levels.}
\end{table}
}

\def\tabrisrefcocogevalsupp#1{
\begin{table}[#1]
\centering
\scalebox{0.7}{
\begin{tabular}{l|p{0.77cm}<{\centering}p{0.77cm}<{\centering}p{0.77cm}<{\centering}p{0.77cm}<{\centering}p{0.77cm}<{\centering}p{0.77cm}<{\centering}|p{0.77cm}<{\centering}p{0.77cm}<{\centering}p{0.77cm}<{\centering}p{0.77cm}<{\centering}p{0.77cm}<{\centering}p{0.77cm}<{\centering}|p{0.77cm}<{\centering}p{0.77cm}<{\centering}p{0.77cm}<{\centering}p{0.77cm}
<{\centering}p{0.77cm}<{\centering}p{0.77cm}
<{\centering}}
\hline
\multirow{2}*{Method} & \multicolumn{6}{c|}{L} & \multicolumn{6}{c|}{M} & \multicolumn{6}{c}{H}  \\ 
\cline{2-19}
~& mIoU & P@50 & P@60 & P@70 & P@80 & P@90 & mIoU & P@50 & P@60 & P@70 & P@80 & P@90 & mIoU & P@50 & P@60 & P@70 & P@80 &
P@90\\
\hline
LAVT \cite{yang2022lavt} & 0.55 & 0.60 & 0.55 & 0.48 & 0.38 & 0.17 & 0.47 & 0.51 & 0.46 & 0.39 & 0.30 & 0.13 & 0.38 & 0.42 & 0.37 & 0.30 & 0.21 & 0.08 \\
PolyFormer \cite{liu2023polyformer} & 0.64 & 0.74 & 0.69 & 0.61 & 0.47 & 0.18 & 0.57 & 0.67 & 0.62 & 0.54 & 0.40 & 0.15 & 0.50 & 0.58 & 0.54 & 0.46 & 0.33 & 0.12\\
X-Decoder \cite{zou2023generalized} & 0.58 & 0.65 & 0.62 & 0.56 & 0.46 & 0.22 & 0.50 & 0.56 & 0.53 & 0.49 & 0.39 & 0.18 & 0.44 & 0.49 & 0.46 & 0.40 & 0.31 & 0.13\\
ETRIS \cite{xu2023bridging} & 0.53 & 0.61 & 0.55 & 0.47 & 0.33 & 0.11 & 0.45 & 0.51 & 0.46 & 0.38 & 0.26 & 0.08 & 0.38 & 0.42 & 0.36 & 0.29 & 0.19 & 0.06 \\
SEEM \cite{zou2024segment}& 0.56 & 0.62 & 0.60 & 0.56 & 0.49 & 0.29 & 0.49 & 0.55 & 0.52 & 0.49 & 0.43 & 0.24 & 0.43 & 0.48 & 0.46 & 0.42 & 0.35 & 0.18\\
\hline
\end{tabular}}
\caption{Robustness of referring image segmentation on RefCOCOg-umd Validation set with low (L), medium (M) and high (H) perturbation levels.}
\end{table}
}

\def\tabrisrefcocogtestasupp#1{
\begin{table}[#1]
\centering
\scalebox{0.7}{
\begin{tabular}{l|p{0.77cm}<{\centering}p{0.77cm}<{\centering}p{0.77cm}<{\centering}p{0.77cm}<{\centering}p{0.77cm}<{\centering}p{0.77cm}<{\centering}|p{0.77cm}<{\centering}p{0.77cm}<{\centering}p{0.77cm}<{\centering}p{0.77cm}<{\centering}p{0.77cm}<{\centering}p{0.77cm}<{\centering}|p{0.77cm}<{\centering}p{0.77cm}<{\centering}p{0.77cm}<{\centering}p{0.77cm}
<{\centering}p{0.77cm}<{\centering}p{0.77cm}
<{\centering}}
\hline
\multirow{2}*{Method} & \multicolumn{6}{c|}{L} & \multicolumn{6}{c|}{M} & \multicolumn{6}{c}{H}  \\ 
\cline{2-19}
~& mIoU & P@50 & P@60 & P@70 & P@80 & P@90 & mIoU & P@50 & P@60 & P@70 & P@80 & P@90 & mIoU & P@50 & P@60 & P@70 & P@80 &
P@90\\
\hline
LAVT \cite{yang2022lavt} & 0.55 & 0.61 & 0.55 & 0.48 & 0.38 & 0.17 & 0.47 & 0.52 & 0.47 & 0.41 & 0.31 & 0.13 & 0.39 & 0.42 & 0.37 & 0.31 & 0.23 & 0.09 \\
PolyFormer \cite{liu2023polyformer}& 0.64 & 0.74 & 0.69 & 0.61 & 0.47 & 0.18 & 0.58 & 0.68 & 0.63 & 0.56 & 0.42 & 0.15 & 0.50 & 0.58 & 0.54 & 0.47 & 0.34 & 0.12\\
X-Decoder \cite{zou2023generalized} & 0.60 & 0.69 & 0.65 & 0.60 & 0.49 & 0.23 & 0.52 & 0.60 & 0.57 & 0.52 & 0.42 & 0.18 & 0.44 & 0.51 & 0.48 & 0.42 & 0.33 & 0.15\\
ETRIS \cite{xu2023bridging} & 0.53 & 0.60 & 0.55 & 0.47 & 0.34 & 0.11 & 0.46 & 0.52 & 0.46 & 0.38 & 0.27 & 0.08 & 0.38 & 0.42 & 0.36 & 0.30 & 0.19 & 0.05\\
SEEM \cite{zou2024segment}& 0.57 & 0.64 & 0.62 & 0.58 & 0.52 & 0.31 & 0.51 & 0.58 & 0.56 & 0.52 & 0.45 & 0.26 & 0.45 & 0.51 & 0.48 & 0.45 & 0.37 & 0.20\\
\hline
\end{tabular}}
\caption{Robustness of referring image segmentation on RefCOCOg-umd Test set with low (L), medium (M) and high (H) perturbation levels.}
\end{table}
}

\def\tabrisrefcocogtestbsupp#1{
\begin{table}[#1]
\centering
\scalebox{0.7}{
\begin{tabular}{l|p{0.77cm}<{\centering}p{0.77cm}<{\centering}p{0.77cm}<{\centering}p{0.77cm}<{\centering}p{0.77cm}<{\centering}p{0.77cm}<{\centering}|p{0.77cm}<{\centering}p{0.77cm}<{\centering}p{0.77cm}<{\centering}p{0.77cm}<{\centering}p{0.77cm}<{\centering}p{0.77cm}<{\centering}|p{0.77cm}<{\centering}p{0.77cm}<{\centering}p{0.77cm}<{\centering}p{0.77cm}
<{\centering}p{0.77cm}<{\centering}p{0.77cm}
<{\centering}}
\hline
\multirow{2}*{Method} & \multicolumn{6}{c|}{L} & \multicolumn{6}{c|}{M} & \multicolumn{6}{c}{H}  \\ 
\cline{2-19}
~& mIoU & P@50 & P@60 & P@70 & P@80 & P@90 & mIoU & P@50 & P@60 & P@70 & P@80 & P@90 & mIoU & P@50 & P@60 & P@70 & P@80 &
P@90\\
\hline
LAVT \cite{yang2022lavt} & 0.69 & 0.78 & 0.74 & 0.68 & 0.56 & 0.27 & 0.57 & 0.64 & 0.60 & 0.53 & 0.42 & 0.18 & 0.46 & 0.51 & 0.46 & 0.40 & 0.29 & 0.12 \\
PolyFormer \cite{liu2023polyformer} & -- & -- & -- & -- & -- & -- & -- & -- & -- & -- & -- & -- & -- & -- & -- & -- & -- & -- \\
X-Decoder \cite{zou2023generalized} & 0.63 & 0.72 & 0.69 & 0.64 & 0.52 & 0.25 & 0.54 & 0.61 & 0.59 & 0.53 & 0.43 & 0.19 & 0.47 & 0.54 & 0.50 & 0.45 & 0.35 & 0.15\\
ETRIS \cite{xu2023bridging}& 0.49 & 0.56 & 0.50 & 0.43 & 0.30 & 0.09 & 0.45 & 0.51 & 0.45 & 0.38 & 0.26 & 0.08 & 0.38 & 0.41 & 0.37 & 0.30 & 0.19 & 0.05\\
SEEM \cite{zou2024segment}& 0.61 & 0.69 & 0.67 & 0.63 & 0.56 & 0.34 & 0.53 & 0.60 & 0.58 & 0.54 & 0.47 & 0.26 & 0.46 & 0.51 & 0.49 & 0.45 & 0.38 & 0.21\\
\hline
\end{tabular}}
\caption{Robustness of referring image segmentation on RefCOCOg-google Validation set with low (L), medium (M) and high (H) perturbation levels.}
\end{table}
}

\def\tabrvosPfull#1{
\begin{table}[#1]
\centering
\scalebox{0.87}{
\begin{tabular}{l|p{0.65cm}<{\centering}p{0.65cm}<{\centering}|p{0.65cm}<{\centering}p{0.65cm}<{\centering}|p{0.65cm}<{\centering}p{0.65cm}<{\centering}|p{0.65cm}<{\centering}p{0.65cm}<{\centering}|p{0.65cm}<{\centering}p{0.65cm}<{\centering}|p{0.65cm}<{\centering}p{0.65cm}<{\centering}|p{0.65cm}<{\centering}p{0.65cm}<{\centering}|p{0.65cm}<{\centering}p{0.65cm}<{\centering}}
\hline
\multirow{2}*{Method} & \multicolumn{10}{c|}{Environment} & \multicolumn{6}{c}{Sensor}  \\
\cline{2-17}
~& \multicolumn{2}{c}{Snow} & \multicolumn{2}{c}{Fog} & \multicolumn{2}{c}{Frost} & \multicolumn{2}{c}{Spatter} & \multicolumn{2}{c}{Bright} & \multicolumn{2}{c}{Defocus} & \multicolumn{2}{c}{Gau. B} & \multicolumn{2}{c}{Motion} \\
\hline
ReferFormer & 42.8 & 46.2 & 45.9 & 50.3 & 41.4 & 45.3 & 45.9 & 49.5 & 48.2 & 53.0 & 41.8 & 45.3 & 42.8 & 46.3 & 41.0 & 44.8 \\
$\text{R}^2$-VOS & 44.3 & 49.4 & 48.2 & 54.1 & 36.7 & 42.6 & 48.0 & 54.7 & 53.8 & 59.6 & 45.8 & 50.1 & 47.6 & 52.2 & 45.7 & 50.3 \\
OnlineRefer & 42.8 & 46.2 & 45.9 & 50.3 & 41.4 & 45.3 & 45.9 & 49.5 & 48.2 & 53.0 & 41.8 & 45.3 & 42.8 & 46.3 & 41.0 & 44.8 \\
SgMg & 52.8 & 56.4 & 54.7 & 59.2 & 48.7 & 53.2 & 55.4 & 59.9 & 58.1 & 63.4 & 53.4 & 57.7 & 55.1 & 59.5 & 52.0 & 57.4 \\
\hline
~ & \multicolumn{12}{c|}{Sensor} & \multicolumn{4}{c}{Transmission} \\
\hline
~ & \multicolumn{2}{c}{Glass} & \multicolumn{2}{c}{Impulse} & \multicolumn{2}{c}{Shot} & \multicolumn{2}{c}{Speckle} & \multicolumn{2}{c}{Contrast} & \multicolumn{2}{c}{Saturate} & \multicolumn{2}{c}{JPEG} & \multicolumn{2}{c}{Pixelate} \\
\hline
ReferFormer & 39.6 & 42.5 & 42.7 & 47.2 & 41.1 & 45.4 & 43.4 & 47.2 & 45.3 & 49.5 & 49.0 & 53.1 & 48.0 & 52.1 & 43.6 & 47.5  \\
$\text{R}^2$-VOS & 43.1 & 48.2 & 41.3 & 48.5 & 40.5 & 46.6 & 41.2 & 47.3 & 46.7 & 52.2 & 54.5 & 60.3 & 52.6 & 58.6 & 48.8 & 54.2 \\
OnlineRefer & 39.6 & 42.5 & 42.7 & 47.2 & 41.1 & 45.4 & 43.4 & 47.2 & 45.3 & 49.5 & 49.0 & 53.1 & 48.0 & 52.1 & 43.6 & 47.5 \\
SgMg & 49.8 & 54.7 & 51.3 & 55.3 & 51.1 & 54.7 & 51.8 & 55.8 & 54.2 & 58.4 & 57.8 & 62.8 & 57.4 & 62.3 & 54.3 & 59.2\\
\hline
\end{tabular}}
\caption{Robustness of referring video object segmentation methods under different textual and visual perturbations on AVS-Multi.}
\end{table}
}

%% file: src/1-intro.tex
\section{Introduction}

Perception systems function as input channels for intelligent systems, analogous to human eyes. \textit{Referring perception}, focuses on identification and contextualization of visual entities by referring multimodal guidance, illustrated in \cref{fig:teaser} (a). It effectively creates a communicative bridge between humans, who issue instructions, and the environment that is subject to perception. Multimodal referring cues used for such identification include textual descriptions, example imagery, or auditory signals corresponding to the target object (\cref{fig:teaser} (b)). Within the domain of referring perception, specific tasks such as referring expression segmentation \cite{LAVT,liu2023polyformer}, audiovisual source localization \cite{gao2023avsegformer,li2024towards,zhou2022avs}, and queryable 3D mapping \cite{jatavallabhula2023conceptfusion}, are essential modules of robotic control \cite{huang2023grounded,ahn2022visually}, autonomous navigation \cite{jatavallabhula2023conceptfusion,liu2024dragon} and planning \cite{hu2023look,sun2022plate}, and human-computer interaction \cite{tziafas2021few}.

Recent advancements in referring perception models \cite{li2023robust,ding2023mevis,LAVT} have been witnessed and deployed on resource-constrained platforms \cite{xiong2023efficientsam,wu2023onlinerefer}. However, existing datasets for model training often comprise delicate images with accurate annotations, a condition rarely met in real-world scenarios. 
In realistic settings, as illustrated in \cref{fig:teaser}, perception systems may face various disturbances, such as environmental noise (buzzing sound from the radio), imprecise human instructions (misspelling), and sensor imperfections (image blur), which can significantly challenge the robustness of these models.
Despite their critical importance for safer real-world application, the impact of such perturbations on model performance are not thoroughly investigated in the literature. Evaluating model robustness is further complicated by the diversity of operational environments; the assessment process is typically labor-intensive and relies heavily on expert judgment. Addressing this gap, this paper introduces the first benchmark in the field to systematically assess the resilience of referring perception models, \ie, trying to obtain a reliable and comprehensive answer to the critical question:  ``\textit{How robust are the current referring perception models under realistic perturbations}?''





\begin{figure}[t]
    \centering
    \includegraphics[width=\textwidth]{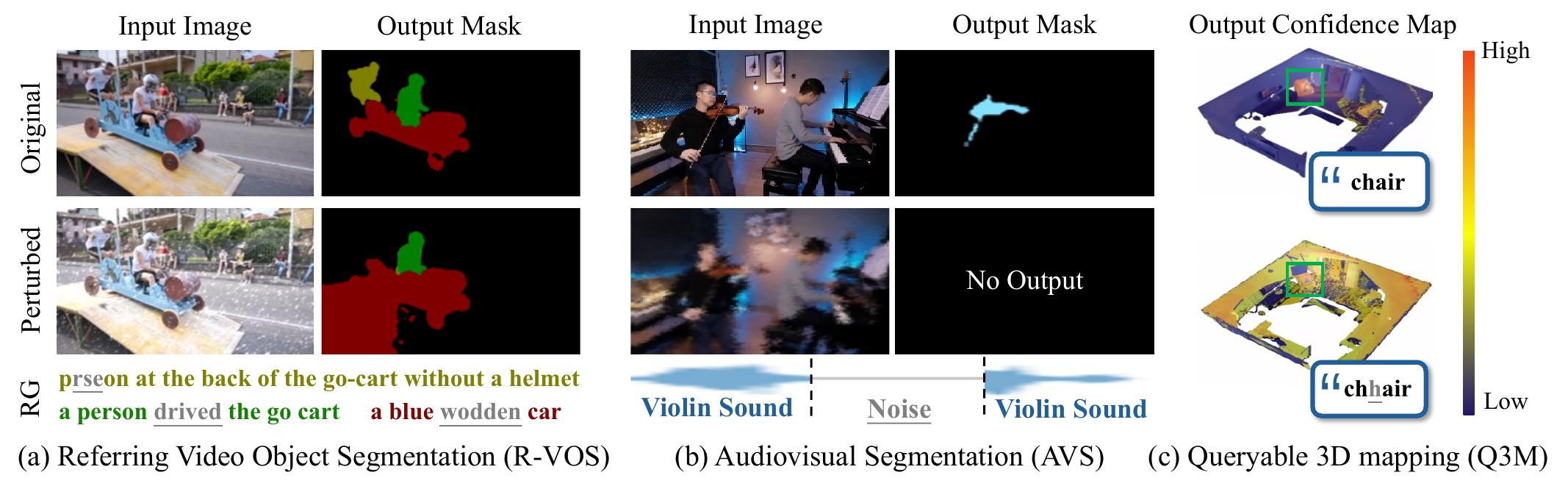}
    \vspace{-0.7cm}
    \caption{
    \textbf{Examples of $\text{R}^2$-Bench.} The ``Original'' row displays the original inputs alongside the outcomes from models of R-VOS  \cite{wu2023onlinerefer}, AVS \cite{li2024towards} and Q3M \cite{jatavallabhula2023conceptfusion}, while the ``Perturbed'' row presents the inputs as synthesized by $\text{R}^2$-Bench and the respective outcomes with the same models. RG: short for Referring Guidance.
    }
    \label{fig:davis}
\end{figure}

Specifically, we introduce the \textbf{R}obust \textbf{R}eferring Benchmark (\textbf{$\text{R}^2$-Bench}), featuring: (1) a human-friendly taxonomy of perturbations for referring perception, categorizing disruptions into source, environment, sensor, and transmission noise, and detailing perturbation types for each modality; (2) a customizable perturbation synthesis toolbox for creating reasonable noise-augmented datasets, enabling robustness evaluation of perception models; and (3) robustness evaluations across five tasks—referring image segmentation (RIS), video object segmentation (VOS), referring video object segmentation (R-VOS), audiovisual segmentation (AVS), and queryable 3D mapping (Q3M)—considering textual, visual, and acoustic reference modalities. We assess over twenty prevalent models in noisy conditions to benchmark their performance and analyze their vulnerabilities. (Examples are illustrated in \cref{fig:davis}.) We hope our benchmarking and analysis can benefit the community by shedding light on the vulnerability of existing models in the face of various perturbations.

Notably, our work advances the evaluation of model robustness by incorporating tests against composite noise types, more closely mirroring real-world scenarios where different noises often coexist. 
The construction of complex noise for realistic robustness assessment is not trivial, as it must avoid non-contextual perturbations, such as the presence of snow in indoor imagery or excessive air absorption effects in indoor audio recordings, which are incongruent with the depicted environment.
To address this complexity, we introduce the \textbf{$\text{R}^2$-Agent}, a novel large language model-based (LLM-based) system that can comprehend natural language instructions provided by humans and autonomously generates test samples with contextual relevant perturbations that align with the deployment environment. For instance, given the human-provided instruction ``outdoor night scene'', the $\text{R}^2$-Agent could probably generate perturbations of ``motion blur'' instead of ``glass blur'' and ``low background noise'' instead of ``room reverberation''.
To improve the agent's abilities of instruction-following and commonsense-reasoning, we employ the multi-agent debating technique \cite{du2023improving,liang2023encouraging,zhao2023competeai}, wherein one agent proposes potential solutions while another verifies their validity. Consequently, the $\text{R}^2$-Agent efficiently automates the identification and integration of mixed perturbations, thus optimizing the process of robustness evaluations tailored to specific domain contexts.



In summary, our major contribution is three-fold:
\begin{itemize}
    \item We introduce the $\text{R}^2$-Bench, a benchmarking framework that includes diverse perturbed data for five commonly encountered referring perception tasks. A comprehensive taxonomy and a corresponding customizable synthesis toolbox are developed to enable robustness assessment in noisy real-world settings. 
    
    \item We propose the $\text{R}^2$-Agent to streamline model evaluations tailored to particular use scenarios based on LLMs.
    This automated program executes the perturbation composition process in accordance with human-provided instructions, thereby allowing for context-specific robustness evaluations.
    
    \item 
    Our systematic experimental investigations delve into the intrinsic characteristics of perturbations, such as their types, severity, dynamics, and correlations. These explorations yield valuable insights into the vulnerability of existing models to disturbances and elucidate the nature of these perturbations.

\end{itemize}

%% file: src/2-related.tex
\section{Related Works}

\paragraph{Textual referring perception.}
Referring image segmentation (RIS) and referring video object segmentation (R-VOS) aim to segment objects in images and video sequences, respectively, based on a linguistic description. Recent RIS methods \cite{LAVT,VLT,liu2023polyformer,zou2023segment,zou2024segment,xu2023bridging,zou2023generalized} have achieved promising results using multimodal transformers. R-VOS \cite{seo2020urvos,wu2023onlinerefer,miao2023spectrum,han2023html,ding2023mevis,tang2023temporal} is more challenging as it requires leveraging both intra-frame and temporal cues. URVOS \cite{seo2020urvos} is the first unified R-VOS framework with cross-modal attention and a memory attention module, significantly improving R-VOS performance. ReferFormer \cite{referformer} employs a linguistic prior in the transformer decoder to focus on the referred object, while MTTR \cite{botach2021mttr} uses a multimodal transformer encoder to fuse linguistic and visual features. $\text{R}^2$-VOS introduces relational cyclic consistency to enhance the robustness of the R-VOS model. Unlike other vision-language tasks \cite{yamazaki2022vlcap, yamazaki2023vltint}, R-VOS needs to construct object-level multimodal semantic consensus in dense visual representations. Relying on 2D models, 3D referring perception is also achievable \cite{yamazaki2023open,jatavallabhula2023conceptfusion}.

\paragraph{Acoustic referring perception.}
Audiovisual segmentation (AVS) \cite{zhou2023audio,liu2024annotation,li2023towards} focuses on segmenting objects that produce sound in a given image frame. This pioneering work by Zhou et al. \cite{zhou2023audio} introduced a method that uses cross-modal attention to identify the sound source. Building on this, Zhou et al. \cite{zhou2023audio} proposed an extended task called audiovisual semantic segmentation (AVSS), which not only segments the sound-producing objects but also classifies them. AVSS is more challenging than AVS due to the complexity of audio semantics. To address this, Zhou et al. \cite{zhou2023audio} utilized the TPAVI module from \cite{zhou2022avs} for audiovisual interaction. Additionally, a recent development by Li et al. \cite{li2023catr} introduced the CATR framework, which features a novel spatial-temporal audio-video fusion block for efficient audio-visual integration. Li et al. \cite{li2024towards} propose a quantization-based semantic decomposition module to handle the complex acoustic environment and enhance the robustness. Sound source localization (SSL) is a related field that aims to identify the visual regions corresponding to sounds. Several SSL methods \cite{arandjelovic2018objects,arandjelovic2017look,cheng2020look,senocak2018learning,li2023panoramic} utilize the correspondence between audio and visual features to locate sounds, often represented as heatmaps. For example, Mo et al. \cite{mo2022closer} employed multi-level audiovisual contrastive learning to effectively pinpoint sound-producing objects. In addition, using speech as referring guidance is also explored in \cite{pan2022wnet,li2023towards}

\paragraph{Visual referring perception.}
Visual referring perception typically relies on visual prompts such as example images, points, bounding boxes, or scribbles. Segment anything model (SAM) \cite{kirillov2023segment} is a powerful foundation model that supports multiple visual prompts. Several follow-up works \cite{xiong2023efficientsam,sam_hq,liu2023grounding} improve SAM in terms of inference cost, segmentation quality, and referring prompts. Beyond that, several works \cite{yu2023inpaint,matte_anything,yang2023track} explore the usage of SAM in downstream tasks. Li et al. \cite{li2023paintseg} built a SAM-like model with stable diffusion. For video-level segmentation tasks \cite{li2023transformer,li2022video,li2022hybrid}, semi-video object segmentation (VOS) \cite{li2024univs,cheng2023putting,yang2022decoupling,yang2021associating,xu2022reliable} aims to segment visual objects across frames given the first frame annotation. Recent offline models \cite{cheng2022xmem,cheng2023putting} focus more on designing long-term information propagation modules to transfer previous image features and corresponding masks to the target frame to predict masks. This helps more precisely identify and track the movement trajectory of objects throughout the entire video sequence (more than 200 frames) but also makes the network architecture heavy \cite{yang2022deaot}.
Due to the slow speed of offline models, online models \cite{zhang2023joint,wu2023segment} come back into focus, aiming to keep the right balance between speed and performance.

%% file: src/3-method.tex
\section{$\text{R}^2$-Bench: Customizable Perturbation Benchmark}
In this section, we present the $\text{R}^2$-Bench for evaluating robustness of referring perception tasks, including a comprehensive taxonomy that categorizes various perturbations and a customizable perturbation synthesis toolbox that can be tailored to generate specific disturbances.

\subsection{Taxonomic Perturbations in Referring Perception}
\label{sec:perturb}
\begin{wrapfigure}{r}{6.4cm}
\vspace{-0.9cm}
\includegraphics[width=0.5\textwidth]{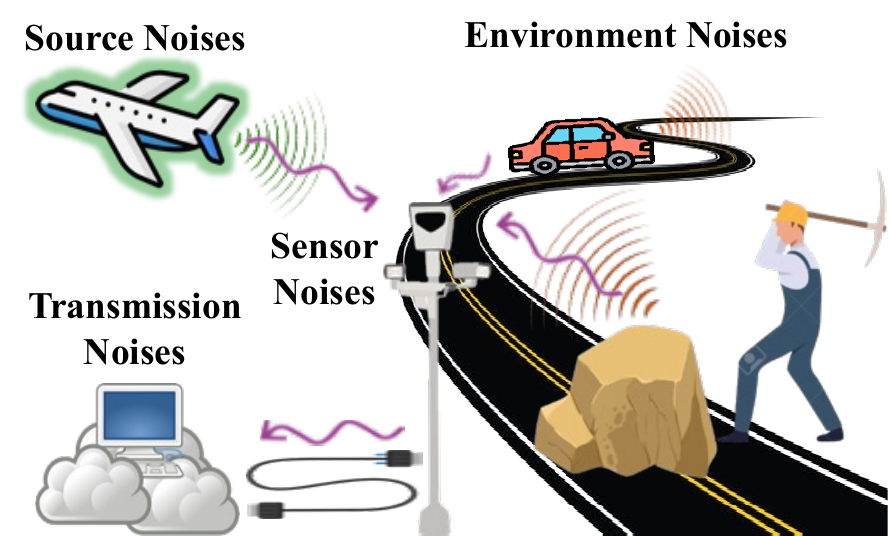}
\vspace{-0.1cm}
    \caption{\textbf{Noise categories based on their origins.} Assuming {\textcolor{forestgreen}{airplane}} as the source of referring guidance, noise from it is categorized as source noise.
    }
    \label{fig:noise_type}
\vspace{-0.9cm}
\end{wrapfigure}
Referring perception tasks typically employ multimodal guidance, spanning the visual, auditory, and textural domains, to ground objects in the visual contexts. Nonetheless, in real-world scenarios, each modality may undergo heterogeneous noises before input into the model. As depicted in \cref{fig:noise_type}, noises can be categorized into four classes based on their origins: source noises, environment noises, sensor noises, and transmission noises. 


\paragraph{Source noises.}
We define source noise as the inherent disturbances introduced spontaneously by the source of the referring guidance. For example, for text-referring tasks, the source noises can manifest as misspellings, grammar errors, or punctuation errors in the textual expression and instruction (human is usually the source of text reference). 

\paragraph{Environment noises.}
We define environment noise as the unintended disturbance introduced by the surrounding environment during the signal capture process. Background sound, room reverberation and 
air absorption for acoustic waveforms are common examples of this type.

\paragraph{Sensor noises.}
We define sensor noise as the unwanted disturbances in the output of sensors that arise from inherent limitations or imperfections. For example, defocus blur, motion blur and impulse noise for visual frames.

\paragraph{Transmission noises.}
We define transmission noise as the signal distortion or loss during the transmission resulting from data compression or package loss. For example, JPEG and MP4 compression for visual and acoustic signals.

\begin{table}[t]
    \centering
    \caption{Noise types considered in $\text{R}^2$-Bench. Details are available in the Appendix.}
    \vspace{-0.3cm}
    \scalebox{0.7}{
    \begin{tabular}{p{1.4cm}|p{3.2cm}|p{3.5cm}|p{5cm}|p{3.6cm}}
    \hline
  \parbox[t]{1.4cm}{\centering\bf Type} & \parbox[t]{3.2cm}{\centering\bf Source} & \parbox[t]{3.5cm}{\centering\bf Environment} & \parbox[t]{5cm}{\centering\bf Sensor} & \parbox[t]{3.6cm}{\centering\bf Transmission}\\
  \hline
  \parbox[t]{1.4cm}{\centering\bf Visual} & \parbox[t]{3.2cm}{\centering -} & \parbox[t]{3.5cm}{\centering snow (SN), fog (FG), frost (FT), spatter (SP), brightness (BR)} & \parbox[t]{5cm}{\centering defocus blur (DB), gaussian blur (GB), motion blur (MB), glass blur (GS), impulse noise (IN), shot noise (ST), speckle noise (SPN), contrast (CT), saturate (SA)} & \parbox[t]{3.6cm}{\centering JPEG compression (JPG), pixelated (PIX)}\\
  \hline
  \parbox[t]{1.4cm}{\centering\bf Acoustic} & \parbox[t]{3.2cm}{\centering amplitude gain (GA)} & \parbox[t]{3.5cm}{\centering background noise (BN), air absorption (AA), room reverberation (RS)} & \parbox[t]{5cm}{\centering gaussian noise (GN), impulse noise (IN), peak filter (PF), time mask (TM), tanh distortion (TD)} & \parbox[t]{3.6cm}{\centering MP3 compression (MP3), lowpass filter (LP), highpass filter (HP)}\\
  \hline
  \parbox[t]{1.4cm}{\centering\bf Textual} & \parbox[t]{3.2cm}{\centering misspelling (MS), mispunctuation (MP), grammar error (GE)} & \parbox[t]{3.5cm}{\centering -} & \parbox[t]{5cm}{\centering character missing (CM)} & \parbox[t]{3.6cm}{\centering -}\\
  \hline
\end{tabular}
}
\label{tab:noise type}
\end{table}

We summarize the supported noises in visual, acoustic and textual modalities in \cref{tab:noise type}. A total of 32 types of noises are considered in this paper. The implementation details of noise functions are available in the Appendix.

\subsection{Customizable Perturbation Synthesis}
We approach the robustness evaluation through two conventional paradigms: (1) general evaluation under universal scenarios and (2) specific evaluation under designated instructions. To facilitate these evaluations, it is essential to generate perturbed datasets from the original clean data samples, denoted as $\mathcal{X}=\{x\}$, by introducing a spectrum of heterogeneous perturbations $\Delta=\{\delta_k \}$.

\paragraph{Perturbation order.} 
As discussed in \cref{sec:perturb}, perturbations arise sequentially according to their origins in real-world contexts. To emulate this, perturbations are applied in a sequence that mirrors the natural order of disruption: source$\rightarrow$environment$\rightarrow$sensor$\rightarrow$transmission.
Within a single category of origin, the order is randomly determined to account for the non-deterministic nature of real-world noise.
To formalize the perturbation process, consider a clean data sample denoted by $x$, the perturbed sample $x^\prime$ undergoes a series of transformations that can be mathematically described as follows:
\begin{equation}
    x^\prime=\delta_{t}\circ\delta_{se}\circ\delta_e\circ\delta_{so}(x),
\end{equation}
where $\circ$ denotes composition operation. $\delta_{so},\delta_{e},\delta_{se},\delta_{t}$ represent generation functions for source, environment, sensor, and transmission noises, respectively.
Given that the introduction of perturbations follows a sequential order, the operations that compose these noise functions are inherently \textit{order-variant}.

\paragraph{Perturbation severity and mode.} In alignment with established methodologies \cite{xu2024customizable,xu2022towards,hendrycks2019benchmarking},
we quantitatively define perturbation severity across discrete levels that align with human perceptual categories, namely, \{low, medium, high\}.
In addition, we investigate perturbation across two main modes based on dynamics: \{static, dynamic\}. Static perturbations remain a consistent level of severity throughout all sensor frames within a sequence, while dynamic perturbations exhibit varying severity and types from frame to frame, closely simulating the fluctuating nature of disturbances in real-world environments, taking the form:
\begin{equation}
    x^\prime=\circ_{k=1}^K\delta_k(x,\lambda_k(t)),
\end{equation}
where $\lambda_k(t)$ denotes the severity of the $k$-th perturbation at time step $t$. 


For the general evaluation under universal scenarios, we construct perturbed datasets for each task by leveraging the original clean benchmarks and all perturbations. Specifically, noisy datasets with low, medium, high and dynamic severity are considered. In each setting, a maximum of two perturbations are considered simultaneously. The detailed data-creating procedure is available in the Appendix. Comprehensive experiments and analyses discussing the impact of perturbations are provided to facilitate future research in \cref{sec:impact}.

Furthermore, we also consider specific evaluations under designated instructions in practice, \eg, virtual phone meetings without background noise. 
We detail $\text{R}^2$-Agent in \cref{sec:agent}.

\begin{figure}[t]
    \centering
    \includegraphics[width=\textwidth]{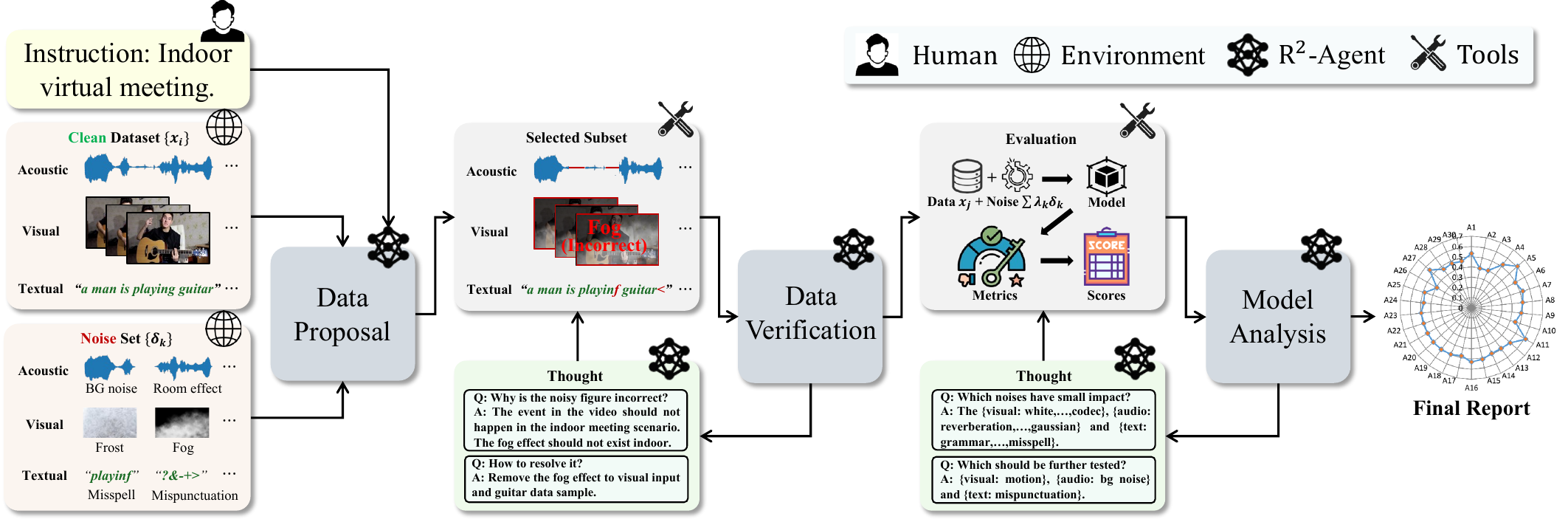}
    \vspace{-0.5cm}
    \caption{\textbf{Overview of $\text{R}^2$-Agent, the automatic evaluation assistant.} Given a human instruction, clean datasets, perturbation functions, and evaluation functions, $\text{R}^2$-Agent first proposes and verifies perturbed test samples that match the given instruction. After that, $\text{R}^2$-Agent evaluates the model using the verified samples and provides a report that articulates the model's vulnerabilities and overall resilience.
    } 
    \label{fig:pipeline}
\end{figure}

\section{$\mathrm{R}^2$-Agent: Automatic Evaluation Assistant}
\label{sec:agent}
With the versatile capabilities of LLMs \cite{achiam2023gpt,team2023gemini}, LLM-based agents \cite{xi2023rise,wang2024describe} are employed to automatically perform various tasks in a human-like manner. In this section, we introduce $\mathrm{R}^2$-Agent, an LLM-based agent that facilitates automatic perturbation synthesis and robustness evaluation guided by human instructions. As depicted in \cref{fig:pipeline}, the $\mathrm{R}^2$-Agent consists of three primary steps: data proposal, data verification, and model analysis. We utilize Gemini-Vision-Pro \cite{team2023gemini} as the LLM component within the $\mathrm{R}^2$-Agent.

\subsection{Evaluation Pipeline}
Given a set of clean data samples $\mathcal{X}=\{x_i\}$, perturbation functions $\Delta=\{\delta_k\}$, evaluation metric functions $\mathcal{E}=\{e_j\}$, and a human instruction $u$, the objective of $\mathrm{R}^2$-Agent is to select an appropriate subset of clean data samples $\mathcal{X}_u$, potential perturbations $\Delta_u$, and suitable evaluation functions $\mathcal{E}_u$ to assess the model based on the given instruction $u$.
A desired human instruction for the $\mathrm{R}^2$-Agent should explicitly specify the scenario (e.g., indoor), the evaluation focus (e.g., segmentation quality), and, if applicable, any specific requirements (e.g., testing for motion blur) during the evaluation process. To enhance the robustness of the $\mathrm{R}^2$-Agent, we introduce a multi-agent debating strategy utilizing an \textit{operator} agent $\Phi$ and a \textit{guardian} agent $\Psi$, each with their own independent memories.

\paragraph{Data proposal.}
To propose data samples that align with the criteria delineated in the human instruction, we employ the \textit{operator} agent $\Phi$ with the text prompts $\texttt{<SEL>}$, 
 $\Phi$ enables the traversal through the combined space of clean samples, pre-defined noise generation functions, and evaluation functions, collectively represented by the tuple $(\mathcal{X}, \Delta, \mathcal{E})$, to select suitable samples. Additionally, we employ an LLM to generate captions $\mathcal{C}$ for the clean samples $\mathcal{X}$. These captions are archived within an accessible memory to enhance the computational efficiency for subsequent stages. The data proposal stage can be formulated as:
\begin{equation}
    \mathcal{C}, \mathcal{X}_u, \mathcal{E}_u, \Delta_u=\Phi(u, \mathcal{X}, \mathcal{E}, \Delta, \texttt{<SEL>}).
\end{equation}

\begin{figure}[t]
    \centering
    \includegraphics[width=\textwidth]{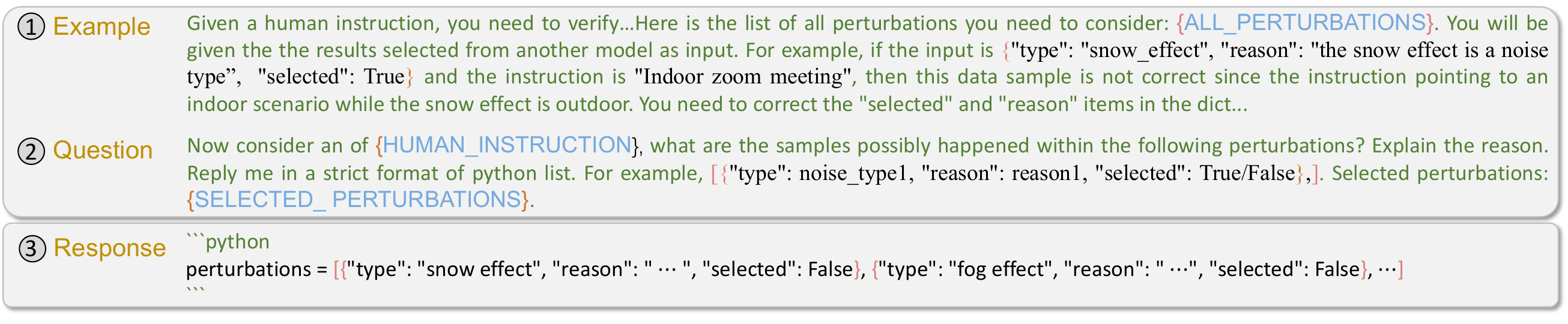}
    \vspace{-0.7cm}
    \caption{\textbf{Chain-of-thought prompting template for data verification.} \ding{172}\&\ding{173}: Following the chain-of-thought spirit, we first give examples to LLM to boost the in-context-learning capability. After that, we ask the LLM to answer a question that is similar to the given example. Specifically, for the data verification task, we ask the LLM to verify the selected samples from the previous iteration, update the results, and explain the reason. The LLM response is instructed to be a Python-format list of dictionaries. \ding{174}: The response from LLM (which follows the desired Python-format list). 
    }
    \vspace{-0.3cm}
    \label{fig:agent}
\end{figure}

\vspace{-0.5cm}
\paragraph{Data verification.}
Although LLMs exhibit notable proficiency in reasoning tasks, 
their susceptibility to generating "hallucinated" content poses a significant challenge to the delivery of accurate responses. 
To mitigate the hallucination effects, we introduce a verification mechanism, designated as the safe guardian $\Psi$, which monitors the data proposal made by the operator $\Phi$ through the utilization of chain-of-thought prompting \cite{wei2022chain}.
We first create the proposed dataset $\mathcal{X}_u^t$ and perturbations $\Delta_u^{t}$. $t$ denotes the iteration number. Then, we construct verification prompts $\texttt{<VER>}$ and employ $\Psi$ to validate the integrity of the data instances. Feedback from this verification is relayed to $\Phi$, which then uses the information to refine and enhance the data proposal.
This iterative process is carried out until a state of convergence or until a maximum iteration number is reached, which is formulated as:
\begin{gather}
    R^t=\Psi(\mathcal{X}_u^t, \Delta_u^{t}, \texttt{<VER>}_\Psi)\notag\\
    \mathcal{X}_u^{t+1}, \Delta_u^{t+1}=\Phi(R^t, \texttt{<VER>}_\Phi),
\end{gather}
where $t$ is the iteration index, $R^t$ is the response of $\Psi$ at $t$ iteration. Note that we consider $\Phi$ and $\Psi$ have separate memories for previous inputs (similar to two separate ChatGPT sessions) and we omit the memorized inputs here for simplicity. 

Specifically, adversarial prompts are leveraged in the data verification step. In $\texttt{<VER>}_\Phi$, we ask the LLM to select conservative choices with high confidence while, for $\texttt{<VER>}_\Psi$, radical choices are acceptable. The adversarial prompts can proactively encourage a debate between two agents thus correcting potential mistakes in the original data proposal. To help understand the data verification step, we demonstrate an example of the data verification process in \cref{fig:agent}.

\paragraph{Model analysis.}
With the selected samples and functions $\mathcal{X}_u, \mathcal{E}_u, \Delta_u$, we instantiate the noisy data $\mathcal{X}_u^\prime$ by applying perturbation functions to the clean data, $\mathcal{X}_u^\prime=\Delta_u(\mathcal{X}_u)$. We then calculate a set of metrics $\mathcal{M}_u=\{m_i\}$ that correspond to $\mathcal{X}_u^\prime$ by applying evaluating functions $\mathcal{E}_u$ to noisy samples. Utilizing the chain-of-though prompt $\texttt{<ANA>}$, we engage an LLM $\Theta$ with empty memory to analyze the metrics and produce an output report $O$ for the model performance as:
\begin{equation}
    O=\Theta(\mathcal{M}_u, \mathcal{X}_u^\prime, \texttt{<ANA>}).
\end{equation}

%% file: src/4-experiments.tex
\section{Experiment}
\subsection{Evaluation Setup and Metrics}
In \cref{tab:bench}, we present a summary of the benchmark tasks, evaluated datasets, and corresponding metrics. Our benchmark encompasses five prevalent tasks, namely, RIS, VOS, R-VOS, AVS and Q3M.

\begin{table}[t]
    \centering
    \caption{Summary of considered tasks, metrics, referring modality, and datasets.}
    \vspace{-0.3cm}
    \scalebox{0.72}{
    \begin{tabular}{c||p{2.6cm}<{\centering}|p{2.7cm}<{\centering}|p{3.5cm}<{\centering}|p{3cm}<{\centering}|p{2.3cm}<{\centering}}
    \hline
    Task & RIS & VOS & R-VOS & AVS & Q3M \\
    \hline
    Ref. Modality & Text & Image & Text & Audio & Text \\
    \hline
    Metrics & mIoU & $\mathcal{G}$, $\mathcal{J}$, $\mathcal{F}$ \cite{KhoRohrSch_ACCV2018} & $\mathcal{J}$, $\mathcal{F}$ \cite{KhoRohrSch_ACCV2018} & $\mathcal{J}$, $\mathcal{F}$ \cite{KhoRohrSch_ACCV2018} & - \\
    \hline
    Dataset & RefCOCO \cite{yu2016refcoco}/+ \cite{yu2016refcoco}/g \cite{mao2016generation}  & DAVIS \cite{KhoRohrSch_ACCV2018}, YTVOS \cite{xu2018youtube} & Ref-DAVIS \cite{KhoRohrSch_ACCV2018}, Ref-YTVOS \cite{seo2020urvos} & AVS-s4 \cite{zhou2023audio}, AVS-ms3 \cite{zhou2023audio} & ScanNet \cite{dai2017scannet}, ICL \cite{handa:etal:ICRA2014}\\
    \hline
    \end{tabular}}
    \vspace{-0.4cm}
    \label{tab:bench}
\end{table}

\paragraph{Evaluation setup.} We conduct systematic evaluations to analyze the impact of perturbations on state-of-the-art mode five tasks. Both perturbations in the referring guidance and visual frames are considered in the benchmarking. We create noisy datasets with low, medium and high noise levels for each task. For video-level tasks, we additionally consider the scenario that noises change over time. We first benchmark the models' general performance with all types of noises and then investigate the impact of noises individually. More information about the dataset creation is available in the Appendix.

\paragraph{Metrics.}
Following the convention \cite{yu2016refcoco,mao2016generation}, we leverage mIoU to evaluate the RIS task. For video-level tasks, the convention is to compute region similarity $\mathcal{J}$ and contour accuracy $\mathcal{F}$ as defined in \cite{pont2017davis}. Specifically, $\mathcal{G}=\frac{\mathcal{J}+\mathcal{F}}{2}$ is also used.  In addition, to better demonstrate the performance degradation against perturbations, we define average performance change (APC) as $    \text{APC}=\sum_i\frac{m_i^{n}-m_i^{c}}{N}$
where $m_i^{n}$ and $m_i^{c}$ denotes the performance of the $i$-th sample of perturbed and clean data respectively. $N$ is the sample number. 
Additional results, evaluated using various metrics and settings, can be found in the Appendix.

\subsection{Performance Benchmarking}

\tabris{t}
\tabvos{t}
\tabrvos{t!}
\paragraph{Referring image segmentation.}
As shown in \cref{tab:ris}, we evaluate the state-of-the-art RIS methods on perturbed RefCOCO/+/g \cite{yu2016refcoco} datasets. We notice that PolyFormer \cite{liu2023polyformer} achieves the best performance in terms of both performance and robustness. Different from other methods that generate pixel-level classification results to represent an object mask, PolyFormer utilizes multi-polygon vertices to represent the mask. We consider the robustness can result from the special sequential prediction and polygon representation which eases the reliance on the per-pixel representation. In addition, we notice that though SEEM \cite{zou2023segment} shows an inferior performance, its robustness to perturbations is promising. We consider that multi-task joint training can account for SEEM's robustness.

\tabavs{t}
\begin{figure}[t]
    \centering
    \includegraphics[width=\textwidth]{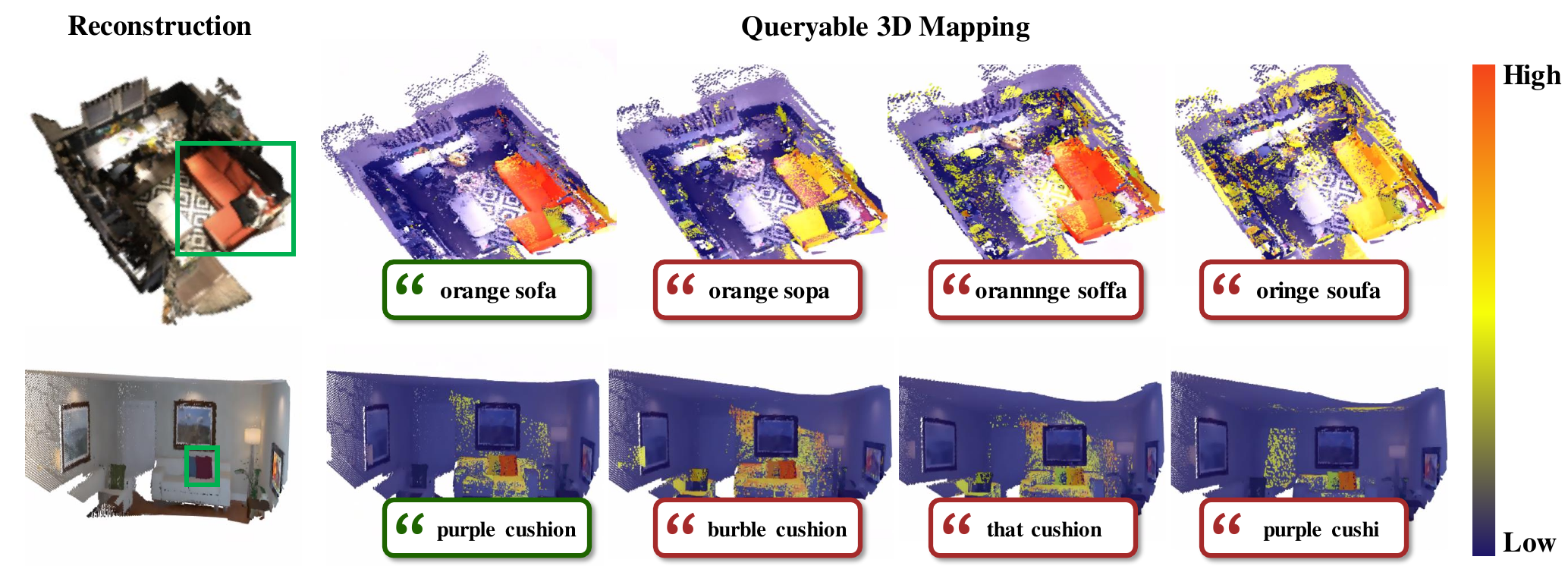}
    \vspace{-0.5cm}
    \caption{Queryable 3D mapping of ConceptFusion \cite{jatavallabhula2023conceptfusion} with inaccurate referring guidance on ScanNet \cite{dai2017scannet} and ICL \cite{handa:etal:ICRA2014} datasets. Due to the absence of publicly available evaluation codes, we did not provide quantitative evaluation for queryable 3D mapping.}
    \vspace{-0.3cm}
    \label{fig:mapping}
\end{figure}
\paragraph{Video object segmentation.}
As depicted in \cref{tab:vos}, we demonstrate the performance of state-of-the-art methods on DAVIS \cite{KhoRohrSch_ACCV2018} and YTVOS \cite{xu2018youtube} datasets. We notice that, for simple scenarios in DAVIS datasets, Cutie \cite{cheng2023putting} shows the best performance. While in the more complex YTVOS dataset, DeAOT \cite{yang2022decoupling} achieves the best. Cutie heavily relies on the pixel- and object-level correspondence across frames which can be disrupted by the perturbations. Differently, DeAOT leverages decoupled visual and ID embeddings which may be the reason for its robustness in noisy scenarios. In addition, we notice that the unseen categories generally have a larger performance drop compared to the seen categories which can impose more challenges in practical deployment in complex scenarios. 

\paragraph{Referring video object segmentation.}
We benchmark the referring video object segmentation task in \cref{tab:rvos}. SgMg \cite{miao2023spectrum} and OnlineFormer \cite{wu2023onlinerefer} achieve the best performance and robustness among methods equipped with video- and image-level backbones respectively. Unlike other methods, OnlineRefer processes the visual features in a frame-by-frame manner which makes it rely less on the pixel-level temporal correspondence that is easily been disrupted by visual perturbations. 

\paragraph{Audio-visual segmentation.}
We show the performance of popular AVS methods in \cref{tab:avs}. We notice that QSD \cite{li2024towards} and CATR \cite{li2023catr} archive promising performance and robustness among all methods. CATR \cite{li2023catr} and AVSegFormer \cite{gao2023avsegformer} are similar methods that leverage acoustic query to query the visual frames with a transformer-based structure. QSD moves one step forward which additionally introduces a quantization operation and a local-global distillation to enhance the robustness of the acoustic representation to adapt to complex scenarios.

\paragraph{Queryable 3D mapping.}
Recent queryable 3D mapping methods \cite{jatavallabhula2023conceptfusion,yamazaki2023open} typically leverage the results from 2D referring models and fuse them to a 3D map. Thereby, the failure of 2D referring perception can directly obstacle the success of 3D tasks. As shown in \cref{fig:mapping}, we visualize the state-of-the-art queryable 3D mapping method, ConceptFusion \cite{jatavallabhula2023conceptfusion}. We notice that with the accurate text query ``orange sofa" ConceptFusion can successfully locate the object while when misspells occur in the query, an obvious performance degradation can be observed. 

\subsection{Perturbation Analysis}
\label{sec:impact}

\begin{figure}[t]
    \centering
    \includegraphics[width=\textwidth]{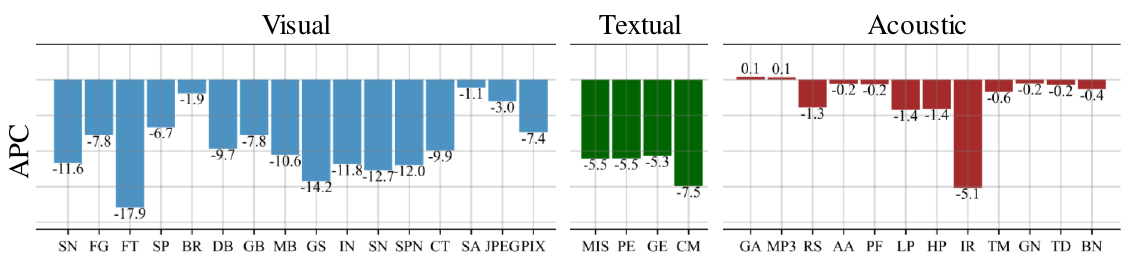}
    \vspace{-0.7cm}
    \caption{\textbf{Average performance changes (APC) for each perturbation type.} 
    \vspace{-0.3cm}
    }
    \label{fig:perf}
\end{figure}
\paragraph{Perturbation type.}
We conduct ablation studies on performance degradation against different perturbation types. \cref{fig:perf} illustrates the aggregated average performance deviation among all benchmarked methods under visual, textual, and acoustic perturbations, respectively. 
For visual perturbations, we notice the frost effect (FT) induced the most substantial performance degradation, which is likely attributable to its extensive occlusion and distortion effects in the whole image. Conversely, color-based perturbations, such as brightness (BR) and saturate (SA), exhibit marginal impact on performance, possibly due to the preservation of shape information. 
For textual perturbations, our analysis indicated that all types of perturbations lead to comparable levels of degradation. The character missing (CA) has a slightly greater impact. 
For acoustic perturbations, we notice that most perturbation types only show a marginal impact on the performance. We consider this is because the target objects in AVS-s4/ms3 are salient objects,
allowing models to rely predominantly on visual cues for object localization, even when the acoustic guidance lacks precision.
Notably, the impulse response perturbation was associated with a significant performance drop, which could stem from the failure in feature extraction due to the impulse signal. Such abnormal acoustic features could potentially disrupt subsequent multimodal interactions.

\paragraph{Perturbation correlation.}

\begin{wrapfigure}{r}{5cm}
\vspace{-0.8cm}
\includegraphics[width=0.4\textwidth]{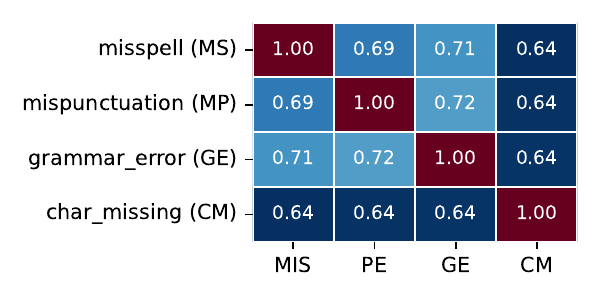}
\vspace{-0.4cm}
\caption{\textbf{Correlation of performance of textual perturbations.}}
\label{fig:txt_corr}
\vspace{-1cm}
\end{wrapfigure}
\begin{figure}[t]
    \centering
    \includegraphics[width=\textwidth]{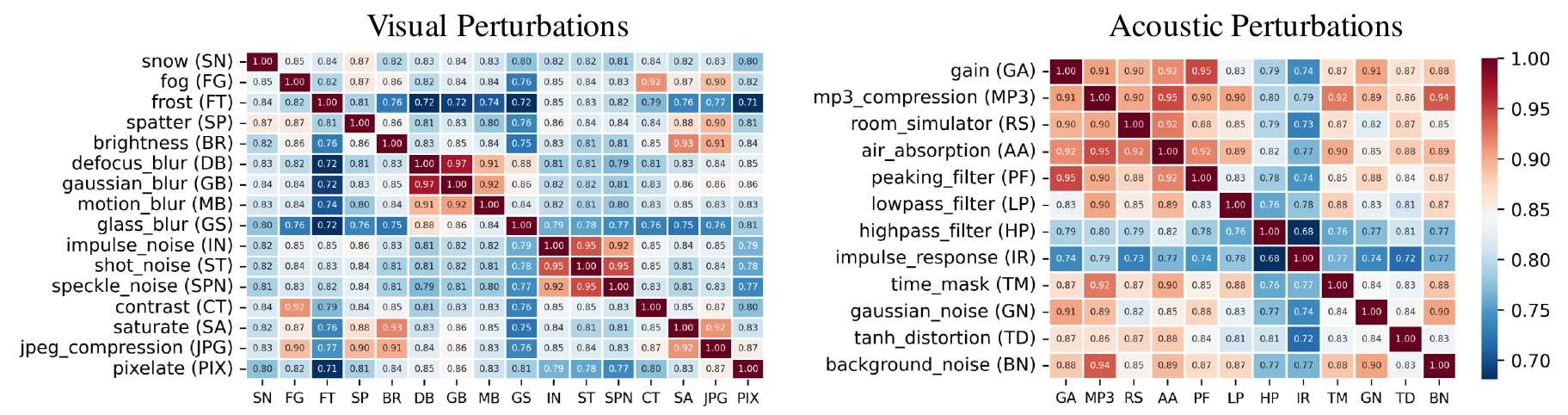}
    \vspace{-0.7cm}
    \caption{\textbf{Correlation matrix of degraded performance with visual and acoustic perturbations.} We concatenate the sample-level performance of all methods on all testing datasets as a feature vector to calculate Pearson product-moment correlation.}
    \vspace{-0.3cm}
    \label{fig:corr}
\end{figure}

To delve deeper into the nature of performance degradation, we computed correlation matrix for each modality, utilizing the Pearson product-moment correlation coefficients calculated from the performance under various perturbation types to populate the matrix. \cref{fig:txt_corr} displays the correlation matrix for textual perturbations. Despite the average performance changes (in \cref{fig:perf}) being similar across different types, the low correlation between the types of performance degradation suggest that each perturbation affects the model in a distinct manner. The correlations for visual perturbations and the ones for acoustic perturbations are presented in \cref{fig:corr} respectively. For visual perturbations, we notice several instances of high correlations. For instance, the group of blurring perturbations—defocus blur (DB), gaussian blur (GB), motion blur (MB), and glass blur (GS)—are highly correlated with one another. For the acoustic perturbations, impulse response (IR) and highpass filter (HF) exhibit different patterns compared to other perturbations, indicating unique effects on performance degradation. We systematically explored the correlations among perturbations and hope the analysis can provide insights for wiser evaluation in the future.

\paragraph{Static v.s. Dynamic.}
\begin{wrapfigure}{r}{5.5cm}
\vspace{-0.9cm}
\includegraphics[width=0.43\textwidth]{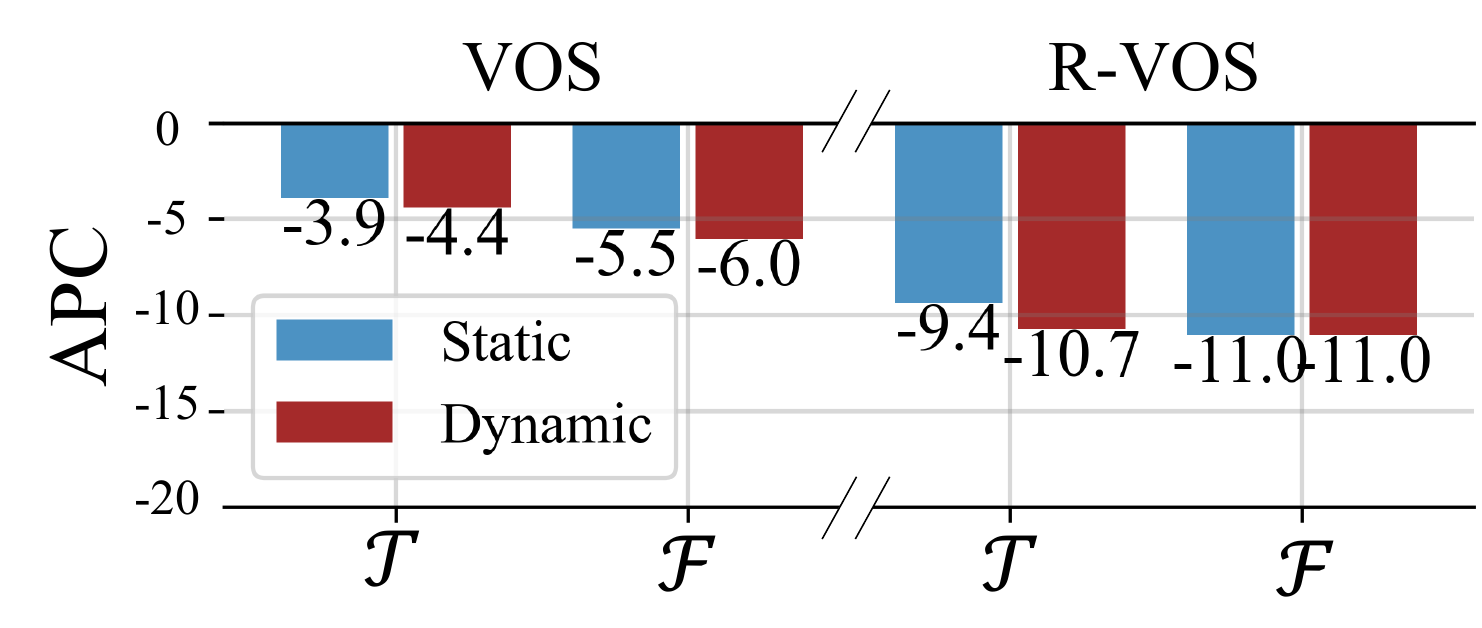}
\vspace{-0.4cm}
\caption{\textbf{Average performance change (APC) under static and dynamic perturbations.}}
\label{fig:dyn}
\vspace{-0.8cm}
\end{wrapfigure}
To analyze the dynamic changing of perturbation types in videos, we conduct an ablation study to explore the impact of perturbation changing across frames (for visual perturbation only). As shown in \cref{fig:dyn}, we calculate the APC across all models on the Ref-DAVIS dataset. We notice the model performance decreases more with dynamic perturbations which can be due to the lack of temporal consistency resulting from various perturbation types. 

\subsection{$\text{R}^2$-Agent: Evaluation Assistant}

\tabdata{r}
\paragraph{User study of data selection.}
To evaluate the data selection performance of $\text{R}^2$-Agent, we manually annotate an evaluation set. Since the instruction can be ambiguous in practice, we measure the alignment between $\text{R}^2$-Agent and human by (1) ACC: accuracy between the human-annotated samples and $\text{R}^2$-Agent predictions and (2) Rate: rating the rationality of the prediction from $\text{R}^2$-Agent (1 if reasonable else 0). As shown in \cref{tab:data}, we ablate on the data verification iteration number. For the metrics function selection, since there are only several hard-coded metrics that can be selected, the accuracy remains perfect while, for data sample and perturbation, we notice a direct performance gain when increasing the iteration number. We notice the performance becomes saturated after 2 iterations, thus we leverage iteration number 2 as the final design choice.
More details can be found in the Appendix.

\begin{figure}[t]
    \centering
    \includegraphics[width=\textwidth]{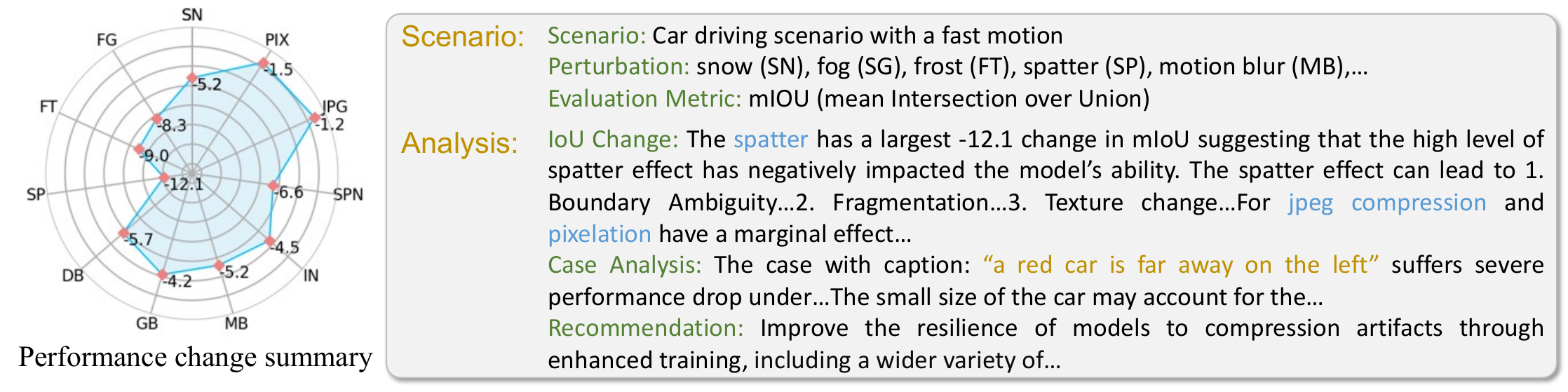}
    \vspace{-0.6cm}
    \caption{\textbf{Example of $\text{R}^2$-Agent report.} The evaluation setting and metrics are fed to $\text{R}^2$-Agent with a set of hard-coded prompts $\texttt{<ANA>}$ to facilitate $\text{R}^2$-Agent generate the human-like report. Beyond the hard-coded analysis, further QA is also supported.}
    \label{fig:report}
\end{figure}

\paragraph{Automatic model analysis.}
With the evaluation setting and corresponding results, $\text{R}^2$-Agent can generate a detailed evaluation report. We demonstrate an evaluation report example in \cref{fig:report}. To make $\text{R}^2$-Agent think in a human-like manner, we leverage a set of chain-of-thought prompts to instruct $\text{R}^2$-Agent to give reasonable and detailed analysis based on the given evaluation results. We notice that $\text{R}^2$-Agent can understand the evaluation metrics and give a human-like analysis which potentially helps to reduce the cost of model evaluation in practice. Furthermore, given the strong dialogue capability of LLMs, further question-answering about the results and report is also feasible. We demonstrate more qualitative results in the Appendix.

%% file: src/5-conclusion.tex
\section{Conclusion}
In this work, we introduced $\text{R}^2$-Bench, a comprehensive benchmark for evaluating the robustness of visual referring perception models against various perturbations. Our contributions include a detailed taxonomy of perturbations, a customizable perturbation synthesis toolbox, systematic perturbation analysis, and $\text{R}^2$-Agent, a novel evaluation assistant based on large language models. Through extensive experiments, we observed and analyzed the intrinsic characteristics of different perturbations to current models and highlighted the importance of robustness in referring perception tasks. A benchmark with 5 popular referring tasks is also provided to facilitate future research. In addition, the data construction and analysis of robustness evaluation can be further simplified with our proposed $\text{R}^2$-Agent. Our findings underscore the necessity for robustness in the deployment of intelligent systems in real-world scenarios, offering a foundation for future advancements in the field. 